\documentclass[authoryear,preprint,review,12pt]{elsarticle}

\usepackage{amssymb}
\usepackage{amsmath,amsfonts,amssymb}
\usepackage{appendix}
\usepackage{xspace}
\usepackage{graphicx}
\usepackage[colorlinks=true, allcolors=blue]{hyperref}
\usepackage[linesnumbered,ruled,vlined]{algorithm2e}
\usepackage{csquotes}
\usepackage{subcaption}
\usepackage{multirow}
\usepackage{comment}
\usepackage{xcolor}
\usepackage{colortbl}
\usepackage{ctable}
\usepackage{tabularx}
\setcitestyle{square}
\setcitestyle{numbers}

\SetCommentSty{mycommfont}
\SetKwInput{KwInput}{Input}                %
\SetKwInput{KwOutput}{Output}              %

\newcommand{\seq}{\mathit{Seq}\xspace}
\newcommand{\units}{u\xspace}
\newcommand{\DIMLEVEL}{\dim_{\mathit{level}}\xspace}
\newcommand{\level}{\mathit{level}\xspace}
\newcommand{\levelmax}{\level^{\max{}}\xspace}
\newcommand{\Neals}{\mathbb{N}\xspace}
\newcommand{\thr}{\mathit{thr}\xspace}
\newcommand{\pluseq}{\mathrel{+}=}

\newcommand{\dataset}{\mathcal{DS}\xspace}
\newcommand{\image}{\mathcal{I}\xspace}
\newcommand{\GT}{GT\xspace}
\newcommand{\nbimages}{\mathit{NbIm}\xspace}
\newcommand{\neuralnetwork}{\Xi\xspace}
\newcommand{\encoder}{\neuralnetwork^\mathit{enc}\xspace}
\newcommand{\classifier}{\neuralnetwork^\mathit{classif}\xspace}

\newcommand{\CreateMatrixOfZeros}{\mathit{CreateMatrixOfZeros}\xspace}
\newcommand{\domain}{\mathcal{D}\xspace}
\newcommand{\sizepatch}{s_\mathit{p}\xspace}
\newcommand{\minimalsizepatch}{\sizepatch^{\mathit{min}}\xspace}
\newcommand{\imagesize}{s_{\mathit{image}}\xspace}

\newcommand{\lx}{\ell_x\xspace}
\newcommand{\ly}{\ell_y\xspace}
\newcommand{\la}{\ell_a\xspace}
\newcommand{\lb}{\ell_b\xspace}
\newcommand{\lxmax}{\ell_x^{\max{}}\xspace}
\newcommand{\lymax}{\ell_y^{\max{}}\xspace}

\newcommand{\lxstar}{\ell_x^{*}\xspace}
\newcommand{\lystar}{\ell_y^{*}\xspace}

\newcommand{\patch}{\mathcal{P}\xspace}

\newcommand{\feature}{\mathcal{F}\xspace}
\newcommand{\numfeature}{n_f\xspace}

\newcommand{\representative}{\mathit{rep}\xspace}
\newcommand{\MAG}{\mathit{MAG}\xspace}

\newcommand{\clustering}{\mathit{Clustering}\xspace}

\newcommand{\argsort}{\mathit{argsort}\xspace}

\newcommand{\activation}{\mathit{Activ}\xspace}

\newcommand{\nbclasses}{\mathit{NbClasses}\xspace}

\newcommand{\sort}{\mathit{sort}\xspace}

\newcommand{\featureimagecarres}{f_p\xspace}
\newcommand{\correlation}{\mathcal{K}\xspace}
\newcommand{\CAOC}{\mathcal{CAOC}\xspace}

\newcommand{\Imp}{Imp\xspace}
\newcommand{\occlusion}{\blacksquare\xspace}
\newcommand{\Reals}{\mathbb{R}\xspace}
\newcommand{\naturals}{\mathbb{N}\xspace}
\newcommand{\initialposition}{\mathit{InitPos}\xspace}
\newcommand{\newposocc}{\mathit{NewPos}^{\occlusion}\xspace}
\newcommand{\importance}{\mathit{Imp}\xspace}
\newcommand{\Position}{\mathit{Position}\xspace}
\newcommand{\indexlocalimage}{i_\mathit{local}\xspace}

\newcommand{\MatriceImportance}{{\mathcal{M}}_{\Imp}\xspace}
\newcommand{\MatriceImportanceAuxiliaire}{{\mathcal{M}}^{\mathit{Aux}}_{\Imp}\xspace}
\newcommand{\MatriceImportanceAuxiliaireDeux}{{\mathcal{M}}^{\mathit{Aux,2}}_{\Imp}\xspace}

\newcommand{\DIMLN}{\dim\_l_n\xspace}
\newcommand{\coords}{\mathit{ListOfCoords}\xspace}
\newcommand{\coordsaux}{\mathit{ListOfAuxiliaryCoords}\xspace}
\newcommand{\PATCHGENERIQUE}[2]{\patch(#1,#2,\sizepatch)\xspace}

\newcommand{\revTwo}[1]{\textcolor{black}{#1}}
\newcommand{\revThree}[1]{\textcolor{black}{#1}}
\newcommand{\revFour}[1]{\textcolor{black}{#1}}

\newcommand{\PD}{\textit{PD}\xspace}

\newcommand{\laurent}[1]{\textcolor{black}{#1}}

\journal{Information Sciences}

\begin{document}

\begin{frontmatter}

\title{Unsupervised discovery of Interpretable Visual
Concepts}

\author[1,2]{Caroline Mazini Rodrigues}

\author[1]{Nicolas Boutry}

\author[2]{Laurent Najman}

\affiliation[1]{organization={Laboratoire de Recherche de l'EPITA -- LRE},%
    addressline={14-16, Rue Voltaire}, 
    city={Le Kremlin-Bicêtre},
    postcode={94270}, 
    country={France}}

\affiliation[2]{organization={Univ Gustave Eiffel, CNRS, LIGM},%
     addressline={5 Boulevard Descartes}, 
    city={Marne-la-Vallée},
    postcode={77454}, 
    country={France}}

\begin{abstract}
 Providing interpretability of deep-learning models to non-experts, while fundamental for a responsible real-world usage, is challenging. Attribution maps from xAI techniques, such as Integrated Gradients, are a typical example of a visualization technique containing a high level of information, but with difficult interpretation. In this paper, we propose two methods, \emph{Maximum Activation Groups Extraction} (MAGE) and \emph{Multiscale Interpretable Visualization} (Ms-IV), to explain the model's decision, enhancing global interpretability. %
 MAGE finds, for a given CNN, combinations of features which, globally, form a \textit{semantic} meaning, that we call \textit{concepts}. We group these similar feature patterns by clustering in \textquote{concepts}, %
 that we visualize through Ms-IV. This last method is inspired by Occlusion and Sensitivity analysis (incorporating causality) and uses a novel metric, called \emph{Class-aware Order Correlation} ($\CAOC$), to globally evaluate the most important image regions according to the model's decision space. We compare our approach to xAI methods such as LIME and Integrated Gradients. Experimental results evince the Ms-IV higher localization and faithfulness values. Finally, qualitative evaluation of combined MAGE and Ms-IV demonstrates humans' ability to agree, based on the visualization, \revThree{with} the decision of clusters' concepts; and, to detect, among a given set of networks, the existence of bias. 
\end{abstract}

\begin{keyword}

explainable artificial intelligence \sep interpretability \sep convolutional neural networks \sep global artificial concepts

\end{keyword}

\end{frontmatter}

\section{Introduction}

\begin{figure}[!ht]
  \centering
  \includegraphics[width=7cm]{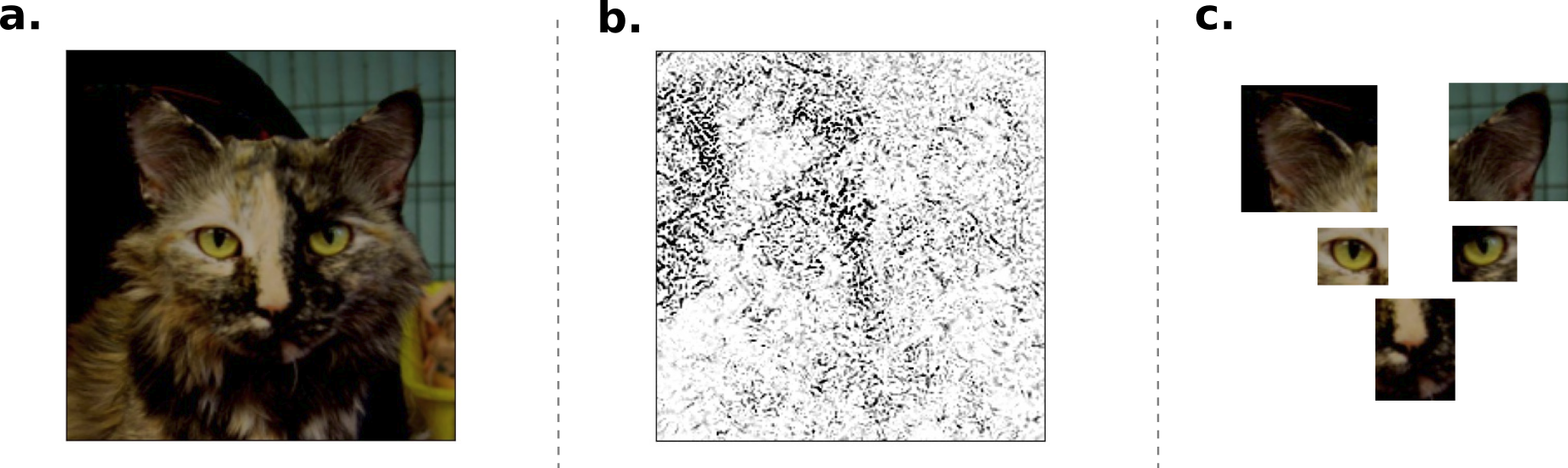}
  \caption{Pixel-level importance is more difficult to interpret than components one. From left to right: an image, its Integrated gradients~\cite{Sundararajan:2017:icml}, and an easier-to-interpret visualization.}
  \label{fig:ex_inter_exp}
\end{figure}

The use of machine learning (ML) in real-world applications increased the need \revTwo{of} explaining decisions to non-computer experts. However, providing model explanations \revTwo{for} isolated features is challenging. Consider the explanation in Fig.~\ref{fig:ex_inter_exp}(b) which is \revThree{annotated via pixel-level importance}: \revThree{one does not directly understand} the model's knowledge. It is not easily \textit{interpretable}.

\textit{Interpretability}, compared to \textit{explainability}, is more subjective as it involves semantics and the idea of how \revFour{humans} understand signals~\cite{adadi:IEEEAccess:2018,guidotti:acm:2018}\revThree{. This process is a} \textit{translation of knowledge} \revTwo{which} depends not only on the information semantics, but also on how it is transmitted and received~\cite{shannon:1948:bell}.

Methods like LIME~\cite{Ribeiro:CKDDM:2016} and KernelSHAP~\cite{lundberg:neurips:2017} propose visualizations based on interpretable components rather than isolated pixels. These components facilitate the human interpretation of how a model understands a sample. However, instead of understanding the complete \revTwo{knowledge of the model}, they explain the behavior of the convolutional neural network (CNN) \revThree{associated with an} individual image.

Works such as TCAVs~\cite{kim:icml:2018} and Explanatory graphs~\cite{zhang:2021:aaai} aim to translate the model's knowledge and \revFour{behavior} given input changes, into interpretable concepts. Apart from increased interpretability, the required supervision can affect how a model is explained. In the case of TCAVs, we need to know the concepts we are testing the model against. In the case of Explanatory graphs, we need to train a time-expensive model to approximate a graph of activation patterns for a set of images.

These \revTwo{supervision} constraints \revFour{may} impact how well a model can be explained, \textit{i.e.}, if the explanation provided is complete enough to represent the model reasoning. To solve this problem, ACE~\cite{ghorbani:2019:neurips} was proposed to use image segments, represented by internal activations, clustered as \textit{concepts}. In this way, TCAV no longer requires user concept supervision. However, as an example-based technique, it depends on how and \revFour{which} images are segmented, \textit{i.e.}, if we do not use images \revTwo{that contain} all concepts, \revTwo{some} concepts could be left out.

We use a similar idea to cluster concepts, but instead of clustering \revTwo{the segment's} activations, we cluster the internal \revTwo{unit's activation} patterns. By doing this, we are able to provide a more global and complete set of concepts.
    
Our methodology tackles three aspects of \revThree{a} CNN's explainability problem: i) we represent the models' knowledge as \textit{completely} and \textit{globally} as possible without supervision;  ii) we obtain explanations based on how humans understand concepts (\revTwo{groups of patterns} with similar \textit{semantics}), and; iii) we provide interpretability to the explanations, enabling the use of intelligent systems by non-experts.

Our main contributions are:

\begin{itemize}

    \setlength\itemsep{.3em}

    \item Maximum Activation Groups Extraction (MAGE), \revTwo{that constructs novel feature-map representations} based on activation patterns \textit{localization} in \textit{multiple} images, instead of the normal activation vectors from individual images;

    \item The Class-aware Order Correlation (CAOC) metric, to determine the impact of \textit{occlusions}, not only in a single image activation but also how, according to the model, this image relates to the others (\textit{dataset relation)};

    \item A Multiscale Interpretable Visualization (Ms-IV), using $\CAOC$ to have an occlusion-based visualization accounting for \textit{dataset relations}; and presenting \revTwo{a}  \textit{hierarchical selection} of \revTwo{the} important image regions (from the complete image to the smallest defined \revTwo{patch size}) to focus human \revThree{attention} on gradually highlighted image parts.

\end{itemize}

Section~\ref{sec:sota} is a \revTwo{review of the literature on} xAI methods, Section~\ref{sec:intuition} presents the intuition of our method, Section~\ref{sec:exp} shows qualitative and human-based experiments, and Section~\ref{sec:conclusion} concludes the paper.
\section{Literature review and Motivation}
\label{sec:sota}

\revTwo{XAI} methods can be broadly categorized as \textit{intrinsic}, \textit{model-specific}, or \textit{post-hoc} and \textit{model-agnostic}.  Examples of \textit{intrinsic} methods are decision trees~\citep{quinlan:ml:1986}, some attention networks~\citep{gu:ieee_med:2021}, and \revTwo{training alongside text} explanations~\citep{park:cvpr:2018}. They are called intrinsic because they do not need an extra mechanism to provide some level of explanation. For these methods, we \revFour{obtain} the explanations directly from the analyzed learning model. 

The \textit{specific-methods} can also cover \textit{intrinsic} models. However, they refer to explanations specifically applied to some determined architectures. For example, the Deconvolution~\citep{zeiler:eccv:2014}, CAM~\citep{zhou:cvpr:2016} and Grad-CAM~\citep{Selvaraju:ICCV:2017} techniques are firstly designed to explain Convolutional Neural Networks. Nevertheless, they are not \textit{intrinsic} but \textit{post-hoc} models, as they are applied to a pre-trained model. %

 According to recent xAI surveys~\citep{crook:2023:arxiv,ali:2023:infoFusion,schwalbe:2023:dataMining}, \revThree{the most commonly referenced methods include}  LIME~\citep{Ribeiro:CKDDM:2016}, SHAP~\citep{lundberg:neurips:2017}, DeconvNet~\citep{zeiler:eccv:2014}, CAM~\citep{zhou:cvpr:2016}, Grad-CAM~\citep{Selvaraju:ICCV:2017}, Guided Grad-CAM~\citep{Selvaraju:ICCV:2017}, DeepLIFT~\citep{shrikumar:icml:2017}, Integrated Gradients~\citep{Sundararajan:2017:icml}, Guided\--Backpropagation~\citep{Springenberg:ICLR:2015}, Saliency maps, and TCAV~\citep{kim:icml:2018}. Their application is disseminated through different domains and is presented in recent research.

The medical domain is one of the biggest applications of xAI techniques. Some literature reviews in this domain mention the use of TCAV as a concept analysis technique~\citep{borys:2023:euRad, priya:2023:icaccs, Sheu:2023:access}; CAM, Grad-CAM, LRP~\citep{bach:plosOne:2015}, SHAP, and LIME as visual-based model explanations~\citep{priya:2023:icaccs, chaddad:2023:sensors}; ACE~\citep{ghorbani:2019:neurips} and Network dissection~\citep{bau:2017:cvpr} \revThree{as a network decomposition technique; and, t-SNE and UMAP~\citep{borys:2023:euRad} as a supplementary visualization technique}. Tim Hulsen~\citep{hulsen:2023:ai} mentions that most of the papers in this area are based on visual \revThree{explanations for} different purposes, such as \revThree{lung} ultrasound~\citep{born:2021:ASci} and breast cancer X-rays~\citep{shen:2021:medical} using CAM; ulcerative colitis colonoscopy using Grad-CAM~\citep{sutton:2022:sciReport}; COVID-19 detection in chest CTs~\citep{lu:2022:intsystems} using Grad-CAM; lung X-ray~\citep{haghanifar:2022:multimedia} using Grad-CAM and LIME, and; chest X-ray images~\citep{Abeyagunasekera:2022:i2ct} using LIME, Integrated Gradients and SHAP.
   
In areas such as network security, models such as LIME, SHAP, and induced decision trees are used to explain the detection of malicious domains~\citep{aslam:2022:interpretable}. There are also applications in forecasting within the manufacturing domain~\citep{chen:2023:springer} using methods such as recursive feature elimination (RFE)~\citep{guyon:2002:machineLearning}, and SHAP.

\revTwo{However, despite their frequent use, we believe that the currently used xAI techniques are not sufficiently interpretable for non-experts to analyze and that they do not fully explain the reasoning of models.} 

\subsection{Model-agnostic methods and interpretability}
\label{sec:inter_sota}

The \textit{model-agnostic} methods are generally \textit{post-hoc} methods and can explain multiple types of architectures. Some examples of \textit{post-hoc} and \textit{model-agnostic} methods \revThree{include not only} Layer-wise Relevance Propagation~\citep{bach:plosOne:2015} and Integrated Gradients~\citep{Sundararajan:2017:icml}, but also LIME~\citep{Ribeiro:CKDDM:2016} and its numerous derivatives~\citep{Ribeiro:CKDDM:2016,Zhou:2021:sigkdd,li:2023:ai}, TreeView~\citep{thiagarajan:arxiv:2016}, and Explanatory graphs~\citep{zhang:aaai:2018}. \revThree{Here, we} describe some methods with a higher level of interpretability and their differences.

\textbf{LIME~\citep{Ribeiro:CKDDM:2016}:} is a \textit{model-agnostic} method which introduces the idea of explaining by using interpretable components. The method decomposes each data sample into human-understandable parts. If a data sample is an image, these decomposed parts can be, for example, superpixels or patches, not necessarily expressed as it is inputted in the model. After we have these parts (or components), LIME measures their importance to a decision. This approach is more human-friendly than showing each individual \revFour{feature's} importance, especially in high-dimensional data. However, despite presenting high interpretability, these explanations are generally local, \textit{i.e.}, they are sample-based or rely on local explanations to explain the model behavior.

\textbf{Explanatory Graphs:} proposed by Zhang~et~al.~\citep{zhang:aaai:2018}, represent a CNN knowledge hierarchy through convolutional layers. Each node in the graph represents candidate patterns of the object's parts, summarizing the knowledge from feature maps. Edges connect nodes from adjacent layers. The method proposes to disentangle object parts from a single filter without ground-truth part annotations. It mines highly activated image patterns from the last convolutional layer (high-level semantics) to the first (simpler structures). This process relies on the complete dataset to optimize the graph of hierarchically connected patterns that best fit network feature maps. 

\textbf{Testing with Concept Activation Vectors (TCAV):} proposed by Kim~et~al.~\citep{kim:icml:2018}, aims to, use a set of low-level features to provide human-friendly, interpretable concepts. In more detail, CAVs' method is part of TCAV, which analyzes how sensitive a model's prediction is to a user's pre-defined concept. The idea is to learn a linear classifier to separate, based on the model internal activations and a given class, the response to the concept's given examples and random \revFour{ones}. TCAV ultimately shows the images most similar to a concept. 

Besides the improvement in interpretability, Explanatory Graphs and TCAVs require \revFour{some level of supervision} to generate the knowledge graphs or to indicate the concepts. \revThree{We also want to obtain a more global network explanation in an unsupervised manner.}

\subsection{Our motivation and similar literature methods}

We propose to extract interpretable visual concepts from a model. \revTwo{Previously}, we mentioned \revFour{some ideas similar to} our proposal: TCAVs and Explanatory Graphs (described in Section~\ref{sec:inter_sota}). \revFour{Additionaly}, a paper proposed by Tan~el~al.~\citep{tan:2022:neuralNets}, has a similar idea: to identify semantic concepts within networks. The authors suggest inducing, during training, neighboring neurons (or feature maps) to exhibit similar activations. The objective is to have an easier interpretation of the activation map visualizations, showing similar regions of activations for similar semantic concepts. \revThree{Their} method, called Locality Guided Neural Network (LGNN), conditions during training, the filters' topology to facilitate manual inspection. However, \revTwo{what distinguishes it from our approach is} that this method focuses on changing the learning algorithm, \textit{i.e.}, it should be used during the training process to change the model. This is not our objective, as we \revThree{aim} to have a general explanation of already trained CNNs.
    
Another work, proposed by Li~et~al.~\citep{li:2018:patternnet}, presents a network, PatternNet, to mine visual patterns that are discriminative and representative. They consider \revFour{that} these patterns should be popular (\textit{representative}), \textit{i.e.}, activated in a considerable number of images from the analyzed class; and unique (\textit{discriminative}) for this class (not appearing in the rest of the classes). However, this is also a supervised task. The authors train PatternNet to find these patterns. Therefore, it is a dataset explanation technique and not a model explanation technique, as we propose.

The method ACE~\citep{ghorbani:2019:neurips} proposes a way to mine concepts, without direct supervision. It uses image segments with different scales to cluster similar patterns, represented by \revTwo{the network activations for each segment}. TCAV is then used to measure the importance of each cluster. We have a similar proposal\revFour{. However}, we aim to represent units as their activation patterns in the complete dataset. In this way, we include a global view of {the network's} behavior while decomposing the network into units with similar concepts.

We believe our methods can fill the gap \revFour{of global model interpretability} by providing unsupervised means of mining semantic patterns inside a pre-trained model, through decomposing the model's \textit{global} knowledge into interpretable concepts.

\section{Proposed methods}

\label{sec:intuition}

Our intuition is: if we can decompose networks' knowledge into different concepts (used for the network decision), we can translate them into human-understandable visualizations (patterns). In this section, we describe these two tasks.

\textbf{\revTwo{Concepts' decomposition:}}
We define \emph{concepts} as combinations of features that form a \emph{semantic meaning}. This generally implies \emph{spatial proximity}. For \revTwo{animal images} (cats \& dogs), a concept can be nearby features that compose a muzzle, ears, or eyes at a given location. Together, these concepts induce \revTwo{the perception of an animal's presence} in the image at this same location.

For digital images, these features are pixels. For the human visual system, the process of grouping pixels into concepts seems \revFour{instinctive}%
, but it is not the same for machines. In the case of artificial neural networks, the patterns that indicate these concepts are learned by internal units during training. 

As previously investigated by other works, we observe different learned patterns inside a CNN by looking into convolutional layers' activations. These patterns can be determined by structures (in the input) that most activate each feature map. It is quite similar to mapping human brain activity. We give a stimulus and look at what lights up in the brain. 

The problem is that CNNs have high-dimensional feature maps, and \revTwo{the} reasoning based on them is humanly infeasible. Our solution is to group similar responses of feature maps' dimensions to provide easier analysis.

\textbf{\revTwo{Concept} discovery through visualization:}
We use a hierarchical visualization strategy to enrich \textit{human understandability}. From a higher to \revFour{a} smaller size of the image substructures (patches), we want to gradually increase attention \revFour{to} the most important parts, from bigger to smaller regions. The idea is similar to a face verification task. The first step is to detect faces in images \revFour{and then to} compare the faces. The complete face is important\revFour{; however}, for verification, facial characteristics such as eyes, nose, and mouth will be more significant. These characteristics are hierarchically linked to the face (inside the face) and during this task, we assign a gradual level of focus, from the face to its specific regions contained within. Similar to this, we expect to facilitate a gradual human attention process, to understand the importance of the main structure and, subsequently, the specific linked characteristics.

The visualization is intended to represent the concepts we previously decomposed. We want to visualize what parts of the original image impact the most for each concept. Using an example: we want to know which \revFour{of}, a dog's muzzle or a dog's eyes, causes the biggest impact in a concept A. In this way, we can discover which concept is A.

We try to capture this \textit{causality} through occlusions. We evaluate the impact of each image region by occluding it and verifying what changed according to a concept A. If the response \revFour{to} concept A changes a lot, the occluded part is important, and \revFour{is} a candidate to explain what is concept A. Different image parts will have different levels of importance. We expect the most important parts to represent the concept.%

However, these occlusions can only be made in each image individually. If we want to account for the \textit{globalism} of a concept (same concept for the majority of images), we need a strategy to include global awareness in the evaluation of the \revFour{concept's} after-occlusion impact.

\revTwo{For our} solution to be globally aware \revTwo{we} evaluate the impacts of the occlusions on the relation between images containing the concept A. Let us consider a pre-defined set of images.
After one image occlusion, we measure how many relationships we have changed in this group. This way, if an image has the concept A, \revTwo{the occlusion of this concept in the image} will change the image's relation to the others (it will have less A than the others). On the other hand, if there is no concept A in this image, even if occlusions change the image's prediction, it should not change the image's relation to the others. 
\newcommand{\FEATUREPATCH}[1]{\feature_{i,\numfeature,#1,\sizepatch}\xspace}

\subsection{Decomposition into concepts: Maximum Activation Groups Extraction}
\label{sec:mage}

\begin{figure}[!ht]
  \centering
  \includegraphics[width=11cm]{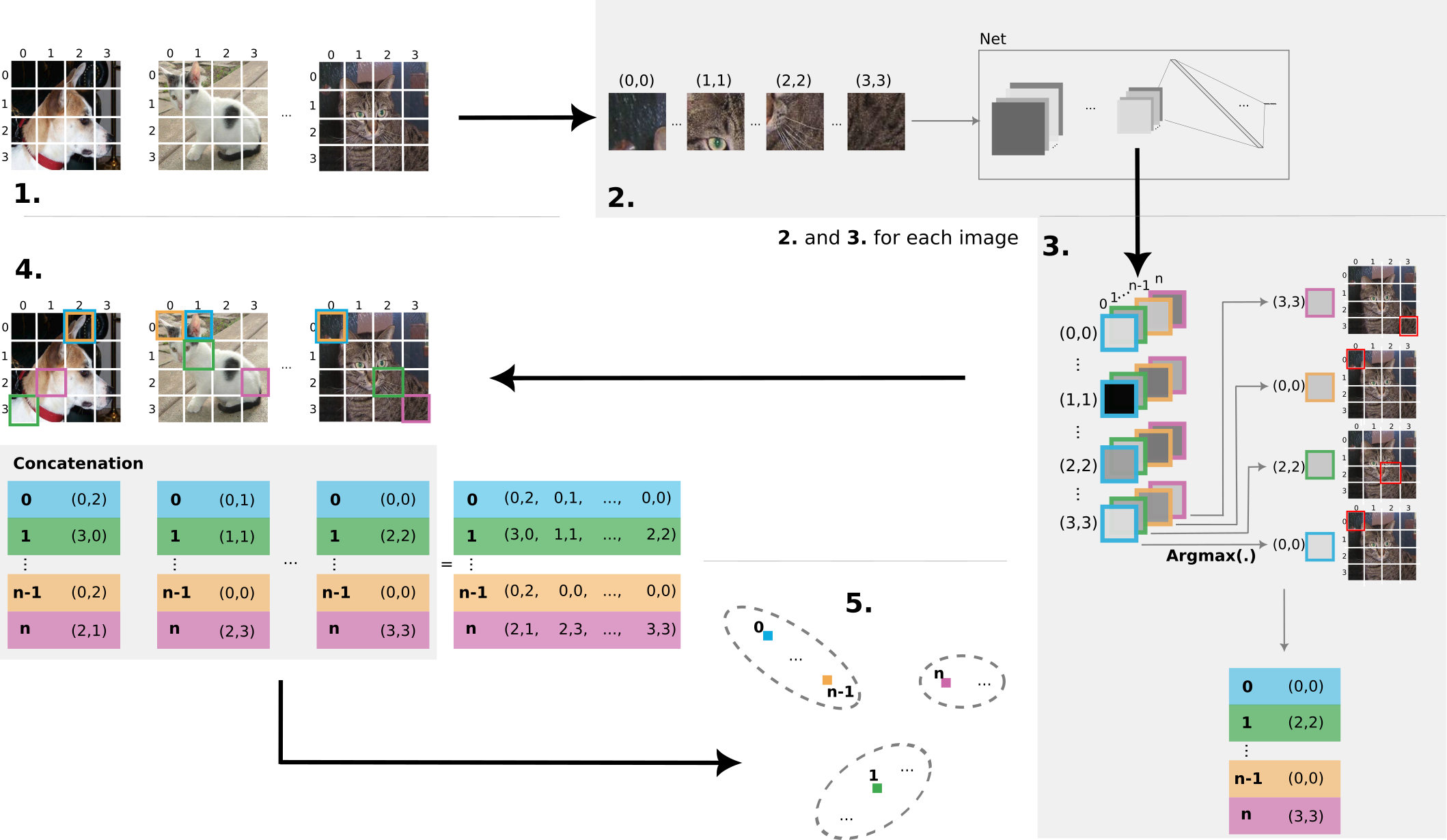}
  \caption{Steps to obtain the \textit{Maximum Activation Groups} (MAGs). We divide dataset images into patches (we perform experiments with different patch sizes to obtain better separation in \textbf{5.}) (\textbf{1.}). We obtain feature maps (from the last convolutional layer) for each patch (\textbf{2.}). We find the corresponding patch with the highest feature map norm by dimension (\textbf{3.}). We concatenate the patches' positions of the highest norms for a set of images to represent each feature map's dimension (\textbf{4.}). We cluster the \revTwo{dimension} representations to obtain the MAGs (\textbf{5.}). More detail in~\ref{ap:mage}.}
  \label{fig:method1}
\end{figure}

The network's knowledge decomposition is made \revTwo{from} five steps (Figure~\ref{fig:method1}).

\subsubsection{Decomposition of input features into patches}
\label{sec:patches}

We decompose the image into non-overlapping patches to evaluate the stimulus of images' subregions to the feature maps. Humans can decompose an image into \textit{semantic pieces} to understand it in its entirety. 

To be able to interpret what CNN-based classifiers \textquote{visualize}, we propose to decompose their feature maps of the highest abstraction level into similar dimensions. Note that we prefer to use the last convolutional layer because it can represent high-level semantic concepts, but we do not use fully-connected layer activations because they lose the \textquote{visual awareness}. To proceed, we start from a CNN like a VGG~\citep{Simonyan:iclr:2015} or a ResNet~\citep{He:cvpr:2016}, that we model using the formula 
\begin{equation}
    \neuralnetwork = \classifier \circ \encoder
    \label{eq:encoder}
\end{equation}
(with $\classifier$ the classifier following the encoder denoted by $\encoder$). The encoder includes the network's layers up to the last convolutional layer. Then, for a given image $\image$, we decompose the image into patches. 

\revTwo{We believe that} this patch decomposition is a key ingredient \revTwo{to better understand} about how the network \textquote{reasons}. \revTwo{To formalize}, let us introduce some notations. The dataset $\dataset = (\image_i,\GT_i)_{i \in [1,\nbimages]}$ used to train $\neuralnetwork$ is made up of $\nbimages$ images $\image_i$ and their class $\GT_i$ (the class number). For a given $i \in [1,\nbimages]$, we denote $\image_i$ as the $i^{th}$ image of $\dataset$. We denote by $\nbimages(c)$ as the number of images of class $c \in [1, \nbclasses]$. For a given class $c$ and for a given $\indexlocalimage \in [1,\nbimages(c)]$, we will denote $\image^c_{\indexlocalimage}$ as the $\indexlocalimage^{th}$ image of class $c \in [1,\nbclasses]$ of $\dataset$.

We can then introduce our formalism\revTwo{:} To decompose the \laurent{domain} $\domain$ of a given image $\image$ into patches of dimension $\sizepatch \times \sizepatch$ (see Figure~~\ref{fig:method1}~\textbf{(1)}), we proceed this way: $\domain = \bigcup_{\lx \in [1,\lxmax], \ly \in [1,\lymax]} \PATCHGENERIQUE{\lx}{\ly},$ with $\lx,\ly,\lxmax,\lymax \in \naturals$, $\lxmax,\lymax$ the number of patches horizontally and vertically (respectively), and $(\lx,\ly)$ the relative coordinates of the patch $\PATCHGENERIQUE{\lx}{\ly}$ described by
\begin{equation}
   \PATCHGENERIQUE{\lx}{\ly} = [1 + (\lx-1) \cdot \sizepatch, \; \lx \cdot \sizepatch] \times [1 + (\ly-1) \cdot \sizepatch, \; \ly \cdot \sizepatch] 
\end{equation}

(we obtain then a partition of $\domain$).
 
\subsubsection{Calculus of feature maps' activation per patch}
\label{sec:activations}

We obtain activations of feature maps by giving each patch to the network --- the response to the stimuli (see Figure~~\ref{fig:method1}~\textbf{(2)}). We use the model described by Equation~\ref{eq:encoder}, in which $\encoder$ includes the \revFour{network layers} up to the last convolutional layer. Since for the image $i$ and for the $\numfeature^{th}$ feature, we have the 2D mapping $\encoder(.,.,\numfeature) : (x,y) \rightarrow \Reals$, and that we will restrict the input image to the patch $\PATCHGENERIQUE{\lx}{\ly}$, we propose to introduce the term $\FEATUREPATCH{(\lx,\ly)} : (x,y) \rightarrow \Reals$, which maps an image patch $\PATCHGENERIQUE{\lx}{\ly}$ into the $\numfeature^{th}$ feature map after a forward pass through $\encoder$, representing the 2D feature map activations for the mentioned patch. Now that we have introduced the formalism to represent feature patches, let us show how we decompose the \textquote{knowledge} of the encoder into \emph{concepts}.

\subsubsection{Identifying important patches for feature map's dimensions}
\label{sec:important_patches}

We want to group feature map dimensions according to their activation patterns. Therefore, we characterize these dimensions by their patterns. This characterization is similar to a brain experiment: if part A of the brain is more activated by emotions than by a task like reading, part A probably knows the concept of emotions. Instead of using emotions and reading, we give the patches to the network. Therefore, we identify the patches that activate a feature map's dimension the most to represent the mentioned dimension. The identification is done by locating the selected patch; in this paper, \textit{\revFour{the process} is based on its position in the original image}.

Let us choose some image $\image_i$ in $\dataset$. We consider in the $\numfeature^{th}$ feature map corresponding to $\image_i$ the one which maximizes the $1$-norm \revFour{as the \emph{reference patch}}. It is then identified by its parameters:
\begin{equation}
    (\lxstar(i,\numfeature),\lystar(i,\numfeature)) = {\arg \max}_{(\lx,\ly)} \left\{ || \FEATUREPATCH{(\lx,\ly)} ||_1 \right\}.
\end{equation}
Intuitively, this position represents the patch where the CNN reacted the most (see Figure~~\ref{fig:method1}~\textbf{(3)}).

\subsubsection{Dimension characterization by the dataset}
\label{sec:representation}

We have limited (local) information if we use only one image to characterize dimensions. Therefore, we incorporate more images in the process. Instead of having the feature map's dimensions characterized by \revTwo{patches of a single image}, we repeat the process \laurent{with} more images to obtain multiple characterizations. We can then define as \textquote{concept} the set of features activated at (almost) the same location for each image of $\dataset$ (see Figure~~\ref{fig:method1}~\textbf{(4)}). In other words, by defining the \emph{representative} of the $\numfeature^{th}$ feature:
\begin{equation}
    \representative(\numfeature) = \left[
\lxstar(1,\numfeature),\lystar(1,\numfeature),\lxstar(2,\numfeature),\dots, \lystar(\nbimages,\numfeature)
\right]^T,
\end{equation}
we obtain a vector in a space (of dimension $2 \nbimages$) which satisfies the property that when two features $\numfeature^1$ and $\numfeature^2$ are physically near to each other in the images of $\dataset$, their representation will be \revFour{near each} other, and \revFour{\textit{vice-versa}}.

\subsubsection{Decomposition of feature space into concepts}
\label{sec:clusters}

We use the \revFour{set} of characterizations of a feature map's dimension to create a feature vector representing it. If two dimensions have similar feature vectors, they activate in similar patches for most images. We consider them the same concept. Therefore, we cluster the feature vectors for all feature map's dimensions to obtain the groups of concepts. This allows us to find the concepts using any clustering algorithm (\laurent{in this paper}, we \revFour{use} $K$-means) to obtain the \emph{Maximal Activation Groups} (MAG):
\begin{equation}
    \{\MAG(k)\}_{k \in [1,K]} = \clustering(K,\left\{\representative(\numfeature) \right\}_{\numfeature}).
\end{equation}
Each term $\MAG(k)$ is what we formally define as a \emph{$\dataset$-relative concept}. They are relative to $K$ and to the clustering algorithm used. Thanks to them, we can understand the \emph{global} behavior of the CNN.

\subsection{Global causality visualization: Multiscale-Interpretable Visualization}
\label{sec:msiv}

\begin{figure}[!ht]
  \centering
  \includegraphics[width=0.80\linewidth]{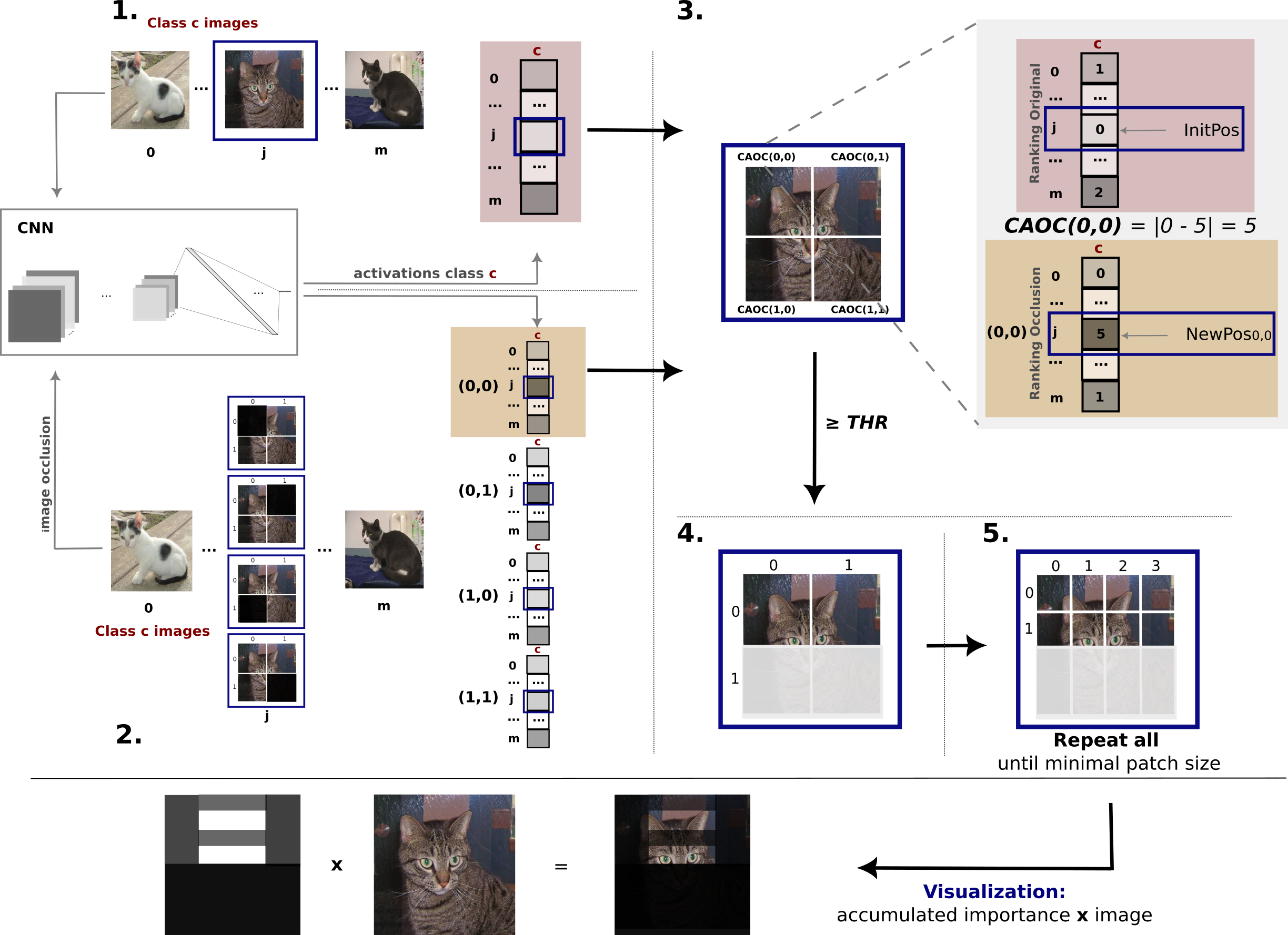}
  \caption{\textit{Multiscale-Interpretable Visualization} (Ms-IV). We obtain the final responses of the network (outputs before Softmax) for a set of images (\textbf{1.}). We occlude one patch (out of 4) from the image we want to visualize, and we calculate the new model's response (\textbf{2.}). We apply \textit{Argmax} to the image's model responses to obtain orders according to a class \textbf{c}. We want two vectors, one ordering all the images, including the original image (to visualize), and; another ordering all images, except the original one (to visualize), but including its occluded version. We obtain the differences in the position of the original and the occluded (\textbf{3.}) to account for the modification in the output space. It is considered the importance of that patch (\textit{Class-aware Order Correlation} ($\CAOC$)).  We do the same to obtain $\CAOC$ for all patches and filter (based on a threshold) the patches that will continue in the next hierarchical visualization level (\textbf{4.}). We reduce the size of patches and repeat the process. We accumulate the importances for all hierarchical levels to, in the end, multiply by the original image and obtain the visualization (\textbf{5.}).  More detail in~\ref{ap:msiv}.}
  \label{fig:method2}
\end{figure}

After \revFour{obtaining} the specific concepts, we follow the inverse path: we look at what a concept represents in each image. We aim to visualize image regions with more impact for the concept $\MAG(k)$ in the model's decision, relying on \textit{human understandability}, \textit{causality}, and \textit{globalism}. 

For the sake of simplicity, let us introduce a new term: we define an \emph{occlusion} of the image $I$ of domain $\domain$ on a patch $\patch \subseteq \domain$ as $\occlusion_\patch(\image) : \domain \rightarrow \Reals$ such that for any $(x,y) \in \domain$, $\occlusion_\patch(\image)(x,y)$ is equal to $\image(x,y)$ when $(x,y) \not \in \patch$, and $0$ otherwise.

We set a visualization threshold ratio $\delta \in \; ]0,1]$, a minimum patch size $\minimalsizepatch \in \Neals^*$ (representing the minimum patch size of a concept), a class $c$, the image $\image^c_{j}$ for visualization, and a concept $k$. The global causality-based visualization is performed in five steps, as shown in Figure~\ref{fig:method2}.

\subsubsection{Original concept output space}
\label{sec:original_space}

Let us define the term \textit{concept output space} in order to describe the image's relation to other dataset images, according to a concept. The \textit{concept output space} is the \revTwo{output matrix of the network}, for a fixed set of images, using only the \revTwo{dimensions of the concepts}, \textit{i.e.} zeroing out all the dimensions belonging to other concepts.

We introduce the notation $\activation(\image;\neuralnetwork)$ which represents the vector of dimension $\nbclasses$ used as input of the softmax layer in the network $\neuralnetwork$ when we input $\image$. %
We set at $0$ the feature activations not relative to concept $k$ in $\neuralnetwork$, leading to a \textquote{new} neural network $\neuralnetwork_k$. The other feature activations are left intact. Then, as illustrated in Figure~\ref{fig:method2}~\textbf{(1)}, we compute the following class-aware matrix,  with the images' activations of class $c$, for the $\MAG(k)$, representing the original \textit{concept output space}:
\begin{equation}
    \mathit{OS}_{\activation}(c,k) =  \left(\activation(\image^c_{\indexlocalimage};\neuralnetwork_k)\right)_{\indexlocalimage \in [1,\nbimages(c)]}
\end{equation}

\subsubsection{Concept output space under input occlusion}
\label{sec:occlusion_space}

Our causal-based visualization employs patch occlusions to identify influential image regions. Depending on the patch size, we create image-specific \textit{concept output spaces} by individually occluding each patch.

We divide the image $\image^c_j$, where $j \in [1,\nbimages(c)]$ that we want to visualize into four patches of the same size $\sizepatch = \frac{\imagesize}{2}$: $\{\PATCHGENERIQUE{\lx}{\ly}\}_{(\lx,\ly) \in \{0,1\}^2}$ (this partitioning assumes that the image size is a multiple of $2$). We perform occlusion on each patch $\PATCHGENERIQUE{\lx}{\ly}$ individually, resulting in a partially occluded image $\occlusion_{\PATCHGENERIQUE{\lx}{\ly}}(\image^c_j)$. That will replace the original image $\image^c_j$ in the original sequence $\left(\image^c_{\indexlocalimage}\right)_{\indexlocalimage \in [1,\nbimages(c)]}$. 

Let us define $\mathit{Occ_{\image^c_j}}(\image^c_{\indexlocalimage}) := \occlusion_{\PATCHGENERIQUE{\lx}{\ly}}(\image_{\indexlocalimage})$ when $\indexlocalimage = j$ and $\image^c_{\indexlocalimage}$ otherwise. Therefore, as presented in Figure~\ref{fig:method2}~\textbf{(2)}, the matrix representing the under-occlusion \textit{concept output space}, of image $\image^c_j$, and patch $\PATCHGENERIQUE{\lx}{\ly}$, is: $\mathit{OS}_{\PATCHGENERIQUE{\lx}{\ly},\activation}(c,k) =  \left(\activation(\mathit{Occ_{\image^c_j}}(\image^c_{\indexlocalimage}));\neuralnetwork_k)\right)_{\indexlocalimage \in [1,\nbimages(c)]}$

\subsubsection{Measuring patch importance}
\label{sec:difference_spaces}

As we have the original and each occluded-patch \textit{concept output space} for an image, we can verify the changes from the original to the under-occlusion spaces. To create our globally aware visualization, we need to verify this impact on the complete space. We propose to use a \textit{ranking-based} approach to measure the difference between the original and an under-occlusion \textit{concept output space}. We name this approach \textit{Class-aware Order Correlation} ($\CAOC$). 

The ranking structure is based on a \textit{target class}, by ordering the points in the \textit{concept output space} from the higher to the lower responses to that class (class-aware). Note that the order of one point depends also on the response of the other points in the space (global awareness). This makes the comparison between the original \textit{concept output space} ranking and the under-occlusion \textit{concept output space} ranking to provide an understanding of how the space changes under a specific occlusion. 

We measure the difference in rankings to determine an importance score. To evaluate the effect of the \revTwo{patch occlusions}, we use rankings. We argsort the values for a class $c$ in the data points on $\mathit{OS}_{\activation}(c,k)$ and $\mathit{OS}_{\PATCHGENERIQUE{\lx}{\ly},\activation}(c,k)$ to obtain the sequence of positions of the class scores, sorted from highest to lowest:
\begin{equation}
    \seq_c = \argsort\left( \mathit{OS}_{\activation}(c,k)_c,\mathit{decreasing}\right)
\end{equation} 
\begin{equation}
    \seq'_c = \argsort\left(\mathit{OS}_{\PATCHGENERIQUE{\lx}{\ly},\activation}(c,k)_c,\mathit{decreasing}\right).
\end{equation}
In practice, in the matrices where each row is a data point (a sample's activations), the activation from class $c$ corresponds to column $c$. Then we compare these sequences. As this sequence of positions are rankings, they can be compared using ranking correlation metrics such as Kendall-tau ($\correlation$). Calculating this correlation measures how much the patch absence impacts the complete space. This way, the score of patch $\PATCHGENERIQUE{\lx}{\ly}$ is obtained by: $\CAOC(\lx, \ly) =\correlation(\seq_c,\seq'_c)$. 

However, as only one image was occluded, we propose to use a simpler calculation of importance. Let us define the image $\image^c_{j}$ position in the original sequence $\seq_c$ as $\initialposition$ and the position of $\occlusion_{\PATCHGENERIQUE{\lx}{\ly}}(\image^c_j)$ in the new sequence $\seq'_c$ as $\newposocc_{\lx,\ly}$. We define the importance of $\PATCHGENERIQUE{\lx}{\ly}$ as: $\CAOC(\lx, \ly) = \left|\initialposition - \newposocc_{\lx,\ly}\right|$ (see Figure~\ref{fig:method2} \textbf{(3)}).

\subsubsection{Choosing important patches}
\label{sec:chosing_patch}

The higher the score, the more important \revFour{the patch}. We use a percentage of the score of the most important patch to define a threshold. Then, we use this threshold to filter the importance of other patches. The visualization consists only of sufficiently important patches. %
We want to consider not only the highest score as important for visualization. However, if we visualize all the patches and their respective scores, we obtain a more confusing and noisy visualization. Therefore, we use a threshold $\thr$ based on $\delta$: $\thr = \max{}\left(\{\CAOC(\lx, \ly)\}_{(\lx,\ly)}\right) \times \delta$. All patches with higher importances will remain in the process (see Figure~\ref{fig:method2} \textbf{(4)}).

\subsubsection{Reducing patch size to repeat process hierarchically}
\label{sec:reducing_patch}

We perform new occlusions of smaller patch sizes in the sufficiently important patches by repeating \revFour{the process from step 2}. We stop reducing \revTwo{the} patch size when we reach a predefined smallest size. We compose the final patch importance by adding up \revTwo{the importance from all patch sizes}. We continue recursively the procedure in the patches satisfying the inequality $\Imp(\lx,\ly) \geq \thr$, returning to step 2. This time, with reduced patch size, while $\sizepatch$ is greater than or equal to $\minimalsizepatch$. 

During this recursive procedure, each position $(x,y) \in \domain$ may have been treated several times. We deduce the \emph{accumulated importance} of a position $(x,y)$ relative to the image $\image_i$ by summing all the computed importance terms where this position was occluded. The final result is called the \emph{accumulated importance matrix} and we denote it $\MatriceImportance$ (see Figure~\ref{fig:method2} \textbf{(5)}). We finally multiply the initial image by $\MatriceImportance$ and we plot it. \laurent{In doing so,} we have highlighted important regions.
\section{Experiments and results}
\label{sec:exp}

Here, we present visualizations of the clusters' dispersion obtained with MAGE and qualitative experiments to visually compare Ms-IV with other methods. We reinforce the results with quantitative evaluation (\textit{Robustness, Faithfulness} and \textit{Localization}). Extra experiments in~\ref{ap:eval}.

To complete the experimentation, we present a concept and bias discovery evaluation with humans. Experiments were performed using two CNN architectures, ResNet-18~\citep{He:cvpr:2016} and VGG16~\citep{Simonyan:iclr:2015}, and datasets: cats vs. dogs~\footnote{https://www.kaggle.com/competitions/dogs-vs-cats-redux-kernels-edition/data} and CUB-200-2011~\citep{he:2019:icsvt} dataset for \textit{Localization} evaluation. Models trained with initial weights from Imagenet, learning rate $1e-7$, cross-entropy loss, the Adam optimizer, and early stop in 20 epochs of non-improving validation loss. \revFour{The code is} at \url{https://github.com/CarolMazini/unsupervised-IVC}.

The cat vs. dog dataset has two classes, dog and cat, with 19,891 images (9,936 dogs and 9,955 cats) in the training set, 5,109 images (2,564 dogs and 2,545 cats) in the validation set, and 12,499 in the test set (without labels). The CUB-200 dataset has 200 classes featuring different bird species. Besides the class annotation (200 labels), it has the \revFour{bird's} bounding box and parts annotation (15 different parts including back, beak, and belly). For more controlled evaluation, we merged 20 different species of warblers and 20 species of sparrows to separate sparrows from warblers. We used the training/validation split provided in the dataset. The two classes each one have 600 images for training and 600 images for validation, totaling 1,200 images in the training set and 1,200 in the validation set.

\subsection{Scatter plot of MAG}

We performed experiments with different patch sizes to generate feature map representations (complete experiments in~\ref{ap:eval}). We chose the one with good final separation of clusters, without being too small (to reduce computation).

\begin{figure}[!ht]
\centering
\begin{tabular}{cc}
\includegraphics[width=5.0cm]{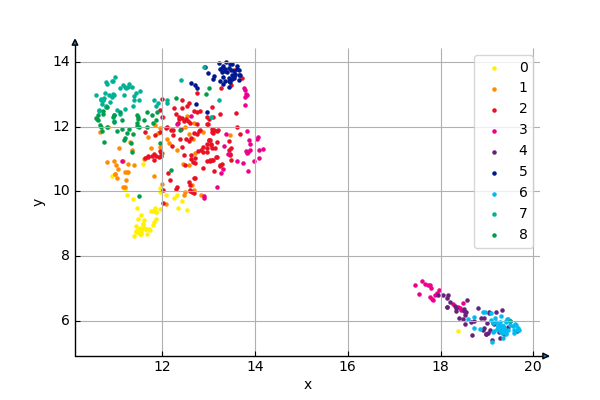} & \includegraphics[width=5.0cm]{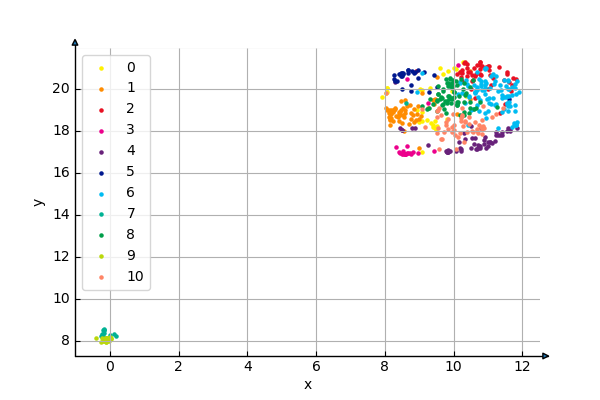}\\
(a) VGG $9$ clusters & (b) ResNet $11$ clusters
\end{tabular}
\caption{Projection of 5120-dimensional representations, from cat vs. dog trained model, to 2D with UMAP. Figures (a) and (b) represent the plots of feature map dimensions from VGG16 and ResNet-18, respectively. Colors represent clusters obtained by K-means. We use $k=9$ for VGG and $11$ for ResNet, chosen by the Elbow curve method using Inertia. }
\label{fig:scatter_clusters}
\end{figure}

In Figure~\ref{fig:scatter_clusters}, \revFour{we show} the dispersion of obtained clusters (high dimensionality reduced to 2D using the Uniform Manifold Approximation and Projection UMAP algorithm~\cite{mcinnes:2018:umap}). To generate the representation, we \revFour{used} a subset of $512$ images (half from each class), $n =4$ (\revTwo{patch} size in representation), and $t=5$ (number of patches per image). To group the concepts, we \revFour{used} K-means, with $k \in [2,25[$ selected by the Elbow curve method and Inertia. The scatter plots, even with the 5120-dimensional representations reduced to 2-dimensional, show the separations of ``clusters of concepts''.

\subsection{Multiscale visualization}

\begin{figure}[!ht]
    \centering
    \begin{subfigure}[t]{0.45\textwidth}
        \centering
\includegraphics[width=5.3cm]{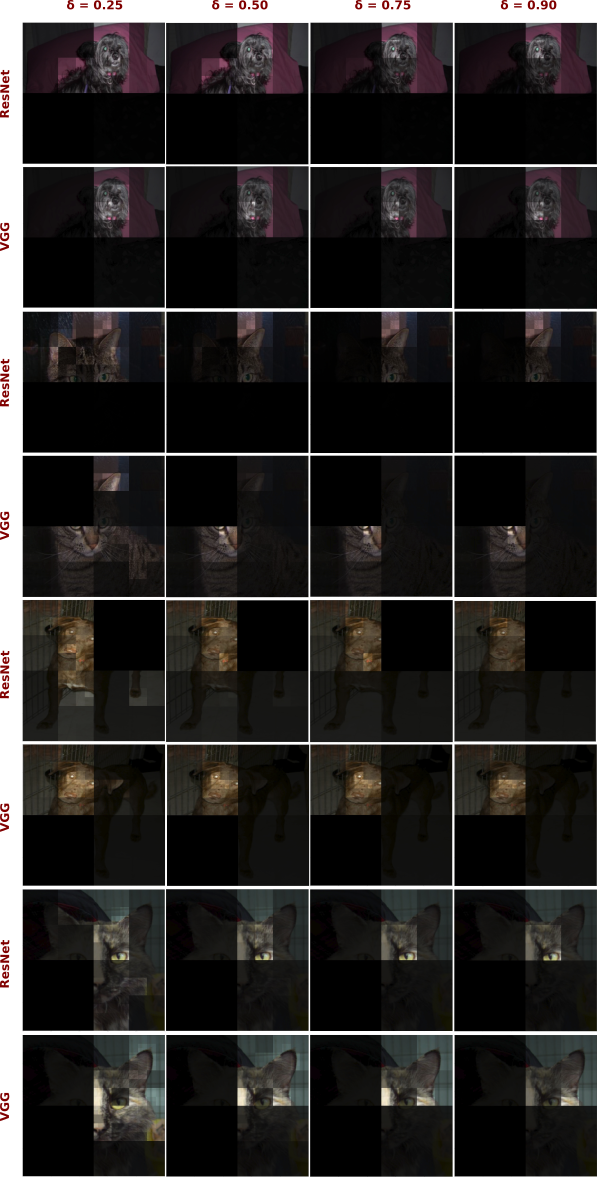}
        \caption{$\CAOC$}
    \end{subfigure}%
    ~
    \begin{subfigure}[t]{0.45\textwidth}
        \centering
\includegraphics[width=5.5cm]{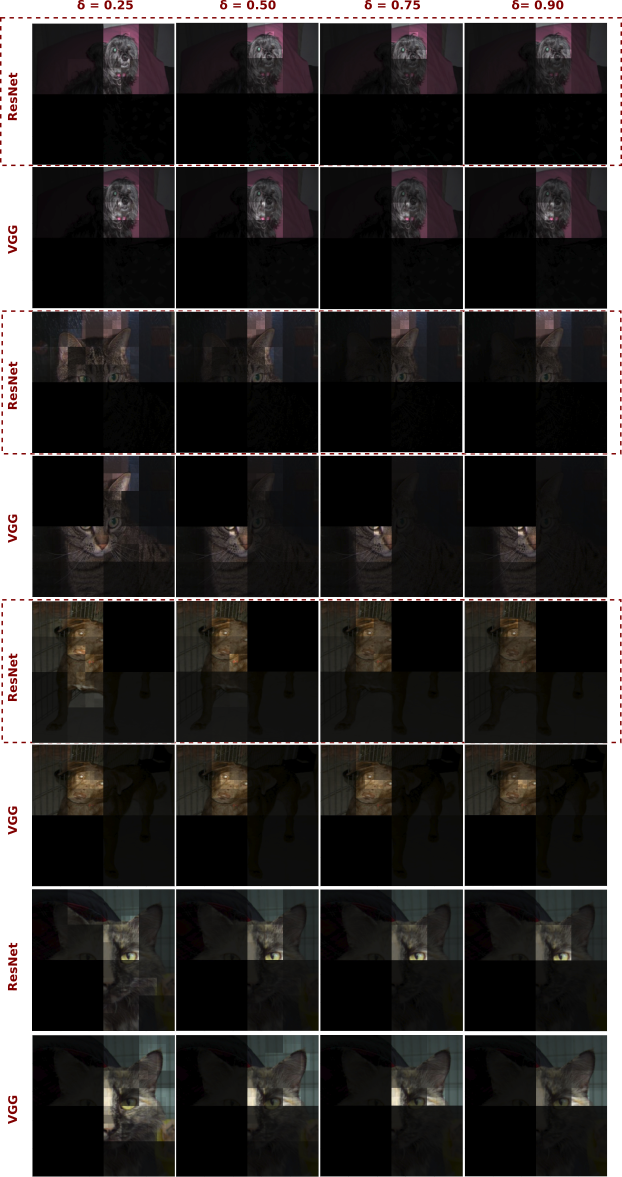}
        \caption{$\PD$s}
    \end{subfigure}
  \caption{Ms-IV visualizations using $\CAOC$ metric to measure patch importance for image samples (two dogs and two cats) using VGG16/ResNet-18 (explanations in the text).
  }
  \label{fig:examples_viz_all_main}
\end{figure}

\begin{figure}[!ht]
  \centering
  \includegraphics[width=11cm]{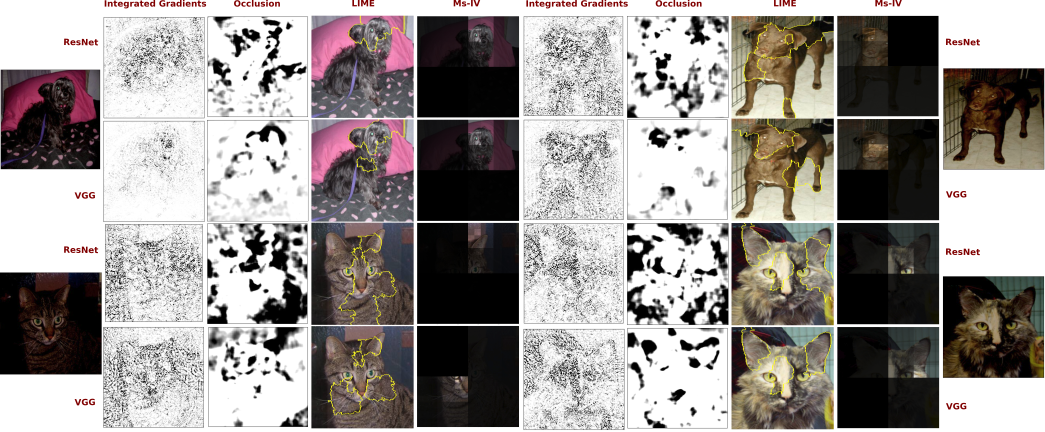}
  \caption{Attribution maps by IG and OC methods versus Ms-IV: IG and OC do not allow us to recognize the shapes of the dog or the cat, whereas Ms-IV, illuminating the initial image, enables us to easily recognize which part of the image is important in the model's decision. Three interpretable components were used for LIME visualizations. The occlusion method uses 7x7 patches, while for Ms-IV we used $thr=0.75$.}
  \label{fig:comp_viz_main}
\end{figure}

We present three visualization experiments: i) visualizations
based on $\thr$ at $ \{0.25, 0.50, 0.75, 0.90\}$ for
changing the acceptance of important patches (calculated by the percentage $\thr$ of the maximum patch importance); ii) visualizations comparing $\CAOC$ and $\PD$s; and, iii) visualizations comparing Ms-IV, Integrated Gradients (IG) and Occlusion (OC).

In Fig.~\ref{fig:examples_viz_all_main}(a) we show results for two values of $\thr$ ($0.25$ and $0.9$) in two image samples and two architectures, using $\CAOC$ to obtain patches contributions. The final value of $\thr$ (last column) shows more focused attention (less bright regions). ResNet18 network focuses more on the animal's eyes than VGG. By switching the metric from $\CAOC$ to $\PD$s (using the same visualization multiscale process), we obtain Fig.~\ref{fig:examples_viz_all_main}(b). For most visualizations, we obtained the same light regions. However, for the examples highlighted by a red dotted rectangle, we see differences: for the dog image ($\thr=0.25$) the dog's paw is highlighted only by $\CAOC$.

These metrics serve for different purposes, $\CAOC$ is sparsity-aware: $\CAOC$ metric will differ from $\PD$s when the model's output space changes density. If the new patch-disturbed image falls in a sparse space region, the patch importance should be smaller, as the region was \textquote{less modified} according to the model. We present in Fig.~\ref{fig:comp_viz_main} the comparison between Ms-IV and two xAI methods: IG and OC.

\subsection{Quantitative evaluation}

Papers such as Bommer~et~al.~\citep{bommer:2023:corr}, analyze the use of metrics such as \textit{Complexity}, \textit{Robustness}, \textit{Faithfulness} and \textit{Localization}, directed towards specific xAI applications. We will discuss their use in our context.

The \textit{complexity} is evaluated based on the number of presented important features. A less complex visualization has fewer important features. Ms-IV uses a $\delta$ parameter to regulate this criterion. Higher $\delta$ highlights fewer patches, facilitating interpretation.

\textit{Robustness} evaluates the impact of adversarial attacks (changing or \revFour{not changing} the classification) on the explanations. For attacks that change the sample's class, we can expect a different explanation (\textbf{Misclassification}); however, for other attacks that do not change the sample's class, we expect the explanations to remain the same (\textbf{Preserved Class}). 

\revFour{For} this evaluation, we \revFour{used} the \textit{Worst Case Evaluation} proposed by \\Huang~et~al.~\citep{huang:2023:iccv}. The method applies a genetic algorithm to find the worst perturbation (adversarial example) for the interpretability of an image explanation. We \revFour{generated} two types of perturbation: one to change the classification but not the explanation, and another to change the explanation but not the classification. We \revFour{perturbed} 30 images (15 per class) by applying a genetic algorithm with 100 iterations, a population size of 100 particles, and a selection of 20 particles for the next iterations (reduced numbers due to computational resource limitations). We \revFour{used} Pearson's correlation to compare the original and perturbed image explanations. We compare Ms-IV to IG and LIME.

\begin{table}[!ht]
\centering
\caption{\textit{Robustness} and \textit{Faithfulness} analysis in the cat vs. dog dataset. (a) \textit{Robustness} values \revFour{were} calculated using Worst Case Evaluation for \textbf{Preserved Class} and \textbf{Misclassification} for three visualization methods: IG, LIME, and Ms-IV. Results \revFour{were} derived using Pearson's correlation between the original image and Worst Case visualizations. Higher values are expected for \textbf{Preserved Class} and lower values for \textbf{Misclassification}. Ms-IV presents a good trade-off between high \textbf{Preserved Class} and low \textbf{Misclassification}. (b) \textit{Faithfulness} analysis based on the percentage of class changes after occlusion (cl.change), decrease of class output value (Decrease), increase of class output value (Increase) using LIME and Ms-IV directed occlusions of 512 images. Additionally, we present the methods' comparison of the biggest variation (absolute output class difference) under occlusion ($>$). Ms-IV presents bigger output variations.}

\begin{minipage}{0.45\textwidth}
\centering
\renewcommand{\arraystretch}{0.65}
\resizebox{.8 \textwidth}{!}{
\begin{tabular}{ccc}
\multicolumn{3}{c}{\cellcolor[HTML]{DAE8FC}\textbf{VGG}}                                                         \\ \hline
                                    & \multicolumn{1}{c|}{\textbf{Preserved Class}} & \textbf{Misclassification} \\ \cline{2-3} 
\multicolumn{1}{c|}{\textbf{IG}}    & \multicolumn{1}{c|}{ \textbf{0.41}}                        &       0.52                     \\
\multicolumn{1}{c|}{\textbf{LIME}}  & \multicolumn{1}{c|}{-0.02}                         &      \textbf{0.08}                      \\
\multicolumn{1}{c|}{\textbf{Ms-IV}} & \multicolumn{1}{c|}{0.34}                &            0.14                \\ \hline
\multicolumn{3}{c}{\cellcolor[HTML]{DAE8FC}\textbf{ResNet}}                                                      \\ \hline
\textbf{}                           & \multicolumn{1}{c|}{\textbf{Preserved Class}} & \textbf{Misclassification} \\ \cline{2-3} 
\multicolumn{1}{c|}{\textbf{IG}}    & \multicolumn{1}{c|}{\textbf{0.43}}                    & 0.50                      \\
\multicolumn{1}{c|}{\textbf{LIME}}  & \multicolumn{1}{c|}{0.078}                    & \textbf{0.01}            \\
\multicolumn{1}{c|}{\textbf{Ms-IV}} & \multicolumn{1}{c|}{0.26}           & 0.29                     
\end{tabular}
}
\subcaption{}
\label{tab:robusteness}
\end{minipage}
~
\begin{minipage}{0.5\textwidth}
\centering
\renewcommand{\arraystretch}{0.65}
\resizebox{.8 \textwidth}{!}{
\begin{tabular}{ccccc}
\multicolumn{5}{c}{\cellcolor[HTML]{DAE8FC}\textbf{VGG}}                                                                                                                           \\ \hline
                                    & \multicolumn{1}{c|}{\textbf{cl. change}} & \multicolumn{1}{c|}{\textbf{Decrease.}} & \multicolumn{1}{c|}{\textbf{Increase.}} & \textbf{\textgreater{}} \\ \cline{2-5} 
\multicolumn{1}{c|}{\textbf{LIME}}  & \multicolumn{1}{c|}{0.03}                & \multicolumn{1}{c|}{0.83}          & \multicolumn{1}{c|}{0.12}          & 0.27                    \\
\multicolumn{1}{c|}{\textbf{Ms-IV}} & \multicolumn{1}{c|}{\textbf{0.08}}       & \multicolumn{1}{c|}{0.80}          & \multicolumn{1}{c|}{0.10}          & \textbf{0.72}           \\ \hline
\multicolumn{5}{c}{\cellcolor[HTML]{DAE8FC}\textbf{ResNet}}                                                                                                                        \\ \hline
\textbf{}                           & \textbf{cl. change}                      & \textbf{Pos.}                      & \textbf{Neg.}                      & \textbf{\textgreater{}} \\ \cline{2-5} 
\multicolumn{1}{c|}{\textbf{LIME}}  & \multicolumn{1}{c|}{0.06}                & \multicolumn{1}{c|}{0.70}          & \multicolumn{1}{c|}{0.23}          & 0.30                    \\
\multicolumn{1}{c|}{\textbf{Ms-IV}} & \multicolumn{1}{c|}{\textbf{0.08}}       & \multicolumn{1}{c|}{0.65}          & \multicolumn{1}{c|}{0.25}          & \textbf{0.69}          
\end{tabular}
}
\subcaption{}
\label{tab:sensitivity}
\end{minipage}
\end{table}

The results in Table~\ref{tab:robusteness} show high \textbf{Preserved class} \textit{robustness} for Ms-IV, closer to the method IG (the best for \textbf{Preserved class} according to Huang~et~al.~\citep{huang:2023:iccv}). Methods such as IG are high resolution (pixel-level importance) and have high robustness when modifications do not change classification results (Table~\ref{tab:robusteness} \textbf{Preserved Class}). However, they lose interpretability (example in Figure~\ref{fig:comp_viz_main}) with noisy visualizations and in robustness when modifications alter the sample's class (Table~\ref{tab:robusteness} \textbf{Misclassification}). LIME significantly improves interpretability, but despite its high robustness for misclassification (Table~\ref{tab:robusteness} \textbf{Misclassification}), it does not have as much robustness as IG for preserved classes (Table~\ref{tab:robusteness} \textbf{Preserved Class}).

\textit{Faithfulness} refers to how much a change in an important feature changes the model's response. For this metric, it is expected to have different outputs (and even different classifications) after perturbations. As we constructed an occlusion-based visualization, we already accounted for \revTwo{perturbation impacts} in the explanations. However, we \revFour{wanted} to verify if the use of occlusion to construct our visualization induced a higher \textit{faithfulness} to the visualization method. To provide a fairer comparison, we compare Ms-IV to LIME, as it also visualizes image regions instead of pixels.

Results in Table~\ref{tab:sensitivity} were obtained using 512 images (256 from each class). We \revFour{extracted} the important region of each image according to both methods: LIME and Ms-IV. We \revFour{used} these regions as masks to occlude the most important image parts. The results in the table represent the percentage of class changes after occlusion (cl.change), decrease of class output value (Decrease), and increase of class output value (Increase). 

Additionally, we \revFour{evaluated} which method disturbs the model's output more in important regions checking, for each image, whether LIME or Ms-IV had the greatest variation (in terms of absolute output class difference) when occluded ($>$). Ms-IV shows a greater number of output variations with $72\%$ of images for the VGG model, and $69\%$ of images for the ResNet model, indicating higher \textit{faithfulness} of the selected important regions to the model.

Ms-IV presents a trade-off between the two methods, IG and LIME (Figure~\ref{tab:robusteness}), and provides interpretable components that, when occluded, have a bigger impact than LIME occlusion components (Table~\ref{tab:sensitivity}).

The criterion \textit{Localization} refers to the ability of a well-trained model to locate the object of interest in the image (of the correct class). For example, in a cat/dog classification problem, if we have a cat for which the model provides the correct answer, the expectation is that the explanation shows the important region inside the cat region (considering an unbiased model).

As in this paper, we aim to decompose and visualize concepts, so the idea of \textit{localization} \revFour{needed} to be adapted. We \revFour{wanted} to evaluate if a MAG (concept cluster) \revFour{could} show the same concept in different images, relying on Ms-IV. To evaluate this, we \revFour{produced} MAGs visualizations using different images and the methods Ms-IV and LIME. Then, we \revFour{used} the human eye to label the displayed areas as different animal parts (a total of 14 different labels). 

As our focus is to be globally aware, we \revFour{considered} that a good-\textit{Locatization} method should show, for different images and the same MAG, the same highlighted animal parts. We also \revFour{evaluated} the original idea of \textit{localization} by calculating the percentage of background highlighted by the visualizations (the lower, the better).  We \revFour{used} 12 individuals to label the 200 image visualizations (reduced amount of visualizations is due to limited human resources).

Table~\ref{tab:localization} presents the results for 10 images of 5 MAGs from both models, VGG and ResNet (a total of 200 visualizations). For each MAG, we show the most frequently labeled animal part within its percentage of appearance in the analyzed images (the higher, the better). Subsequently, we show the background percentage in each MAG. MS-IV \revFour{has} the best conventional \textit{localization} rates (highlighting fewer background regions) and \revFour{presents} higher percentages of the same concept for each MAG, especially for VGG. 

To reinforce the comparison, we also \revFour{performed} a localization experiment with an already parts-labeled dataset, CUB-200-2011~\citep{he:2019:icsvt}. The dataset has 200 bird classes with 60 images each. We \revFour{used} 20 classes of Warblers together and 20 classes of Sparrows together, to compose a binary classification problem. We \revFour{trained} a VGG16 and a ResNet-18 model for this problem, obtaining a validation accuracy higher than $95\%$. There \revFour{were} 16 coordinates of annotated parts per image:  \textit{back, beak, belly, breast, crown, forehead, left eye, left leg, left-wing, nape, right eye, right leg, right-wing, tail, throat}.

First, we \revFour{found} the MAGs for both models. For these bird classification models: VGG \revFour{had} 12 concepts and ResNet \revFour{had} 9 concepts. Second, we \revFour{found} the 100 most activated images for each MAG (each concept). Third, we \revFour{generated} the visualization, LIME and Ms-IV, for each isolated MAG for its 100 most activated images. The idea is, to use the most activated images, to explicitly visualize the concepts. Finally, we \revFour{calculated} the centroid of the highlighted regions in the visualizations and \revFour{compared} it with the coordinates of the parts. We \revFour{considered} the one with the closest coordinate as the visualized part. If the visualization centroid \revFour{was} outside the \revFour{bird's} bounding box, we \revFour{considered} it a background highlight. Tables~\ref{tab:cub_dataset_localization_summary}~and~\ref{tab:cub_dataset_localization} present the results for both models using LIME and Ms-IV.

\begin{table}
\caption{\textit{Localization} for cat vs. dog and CUB datasets. (a) Cat vs. dog dataset considering 5 MAG's concepts (10 images each) for VGG and ResNet evaluated in two ways: Conventional \textit{localization} -- the percentage of background considered important (the lower the better); and concept \textit{localization} -- the quantity of the same concept in each cluster (the higher, the better). Ms-IV presents better \textit{localization} according to 200 visualizations labeled by 12 individuals. For CUB dataset trained models (Warblers vs. Sparrows task): (b) percent of background \textit{localization} and the top2 most-found concepts per MAG; and, (c) concepts \textit{localization} for each MAG. On average, Ms-IV provides better results.}
\begin{minipage}{0.6\textwidth}
\centering
\renewcommand{\arraystretch}{0.65}
\resizebox{.75 \textwidth}{!}{
\begin{tabular}{cccccc}
\multicolumn{6}{c}{\cellcolor[HTML]{DAE8FC}\textbf{VGG}}                                                                                                                      \\ \hline
\textbf{}                           & \multicolumn{1}{c|}{\textbf{0}}                                                              & \multicolumn{1}{c|}{\textbf{1}}                                                                & \multicolumn{1}{c|}{\textbf{2}}                                                           & \multicolumn{1}{c|}{\textbf{3}}                                                        & \textbf{4}                                                               \\ \cline{2-6} 
\multicolumn{1}{c|}{\textbf{LIME}}  & \multicolumn{1}{c|}{\begin{tabular}[c]{@{}c@{}}0.3 faces /\\ 0.2 bellow eyes \end{tabular}}             & \multicolumn{1}{c|}{\begin{tabular}[c]{@{}c@{}}0.5 eyes' \\ region\end{tabular}}               & \multicolumn{1}{c|}{\begin{tabular}[c]{@{}c@{}}0.5 eyes' \\ region\end{tabular}}          & \multicolumn{1}{c|}{\begin{tabular}[c]{@{}c@{}}0.3 chest /\\  0.4 muzzle\end{tabular}} & 0.3 mouth                                                                \\
\multicolumn{1}{c|}{\textbf{Ms-IV}} & \multicolumn{1}{c|}{\textbf{0.4 eyes}}                                                                & \multicolumn{1}{c|}{\textbf{\begin{tabular}[c]{@{}c@{}}0.6 eyes' \\ region\end{tabular}}}      & \multicolumn{1}{c|}{\textbf{\begin{tabular}[c]{@{}c@{}}0.6 eyes' \\ region\end{tabular}}} & \multicolumn{1}{c|}{\textbf{0.7 muzzle}}                                               & \textbf{\begin{tabular}[c]{@{}c@{}}0.4 eyes /\\  0.2 mouth\end{tabular}} \\ \cline{2-6} 
\multicolumn{6}{c}{\textbf{\% background}}                                                                                                                                                                                                       \\ \cline{2-6} 
\multicolumn{1}{c|}{\textbf{LIME}}  & \multicolumn{1}{c|}{0.2}                                                                     & \multicolumn{1}{c|}{0.3}                                                                       & \multicolumn{1}{c|}{0.3}                                                                  & \multicolumn{1}{c|}{0.0}                                                               & 0.0                                                                      \\
\multicolumn{1}{c|}{\textbf{Ms-IV}} & \multicolumn{1}{c|}{\textbf{0.1}}                                                            & \multicolumn{1}{c|}{\textbf{0.2}}                                                                       & \multicolumn{1}{c|}{\textbf{0.2}}                                                         & \multicolumn{1}{c|}{0.0}                                                               & 0.0                                                                      \\ \hline
\multicolumn{6}{c}{\cellcolor[HTML]{DAE8FC}\textbf{ResNet}}                                                                                                                                                                                   \\ \hline
\textbf{}                           & \multicolumn{1}{c|}{\textbf{0}}                                                              & \multicolumn{1}{c|}{\textbf{1}}                                                                & \multicolumn{1}{c|}{\textbf{2}}                                                           & \multicolumn{1}{c|}{\textbf{3}}                                                        & \textbf{4}                                                               \\ \cline{2-6} 
\multicolumn{1}{c|}{\textbf{LIME}}  & \multicolumn{1}{c|}{\begin{tabular}[c]{@{}c@{}}0.3 eyes \\ and muzzle\end{tabular}} & \multicolumn{1}{c|}{\begin{tabular}[c]{@{}c@{}}0.4 eyes'\\  region\end{tabular}}               & \multicolumn{1}{c|}{\begin{tabular}[c]{@{}c@{}}0.4 eyes'\\  region\end{tabular}}          & \multicolumn{1}{c|}{\begin{tabular}[c]{@{}c@{}}0.2 muzzle /\\  0.2 fur\end{tabular}}   & \begin{tabular}[c]{@{}c@{}}0.3 eyes'\\  region\end{tabular}              \\
\multicolumn{1}{c|}{\textbf{Ms-IV}} & \multicolumn{1}{c|}{\begin{tabular}[c]{@{}c@{}}0.3 eyes \\  and forehead\end{tabular}}     & \multicolumn{1}{c|}{\textbf{\begin{tabular}[c]{@{}c@{}}0.3 eyes /\\  0.4 muzzle\end{tabular}}} & \multicolumn{1}{c|}{\textbf{\begin{tabular}[c]{@{}c@{}}0.6 eyes'\\  region\end{tabular}}}          & \multicolumn{1}{c|}{\textbf{0.7 eyes}}                                                 & \textbf{\begin{tabular}[c]{@{}c@{}}0.3 eyes'  region \\/ 0.2 muzzle\end{tabular}}     \\ \cline{2-6} 
\multicolumn{6}{c}{\textbf{\% background}}                                                                                                                                             \\ \cline{2-6} 
\multicolumn{1}{c|}{\textbf{LIME}}  & \multicolumn{1}{c|}{0.4}                                                                     & \multicolumn{1}{c|}{0.5}                                                                       & \multicolumn{1}{c|}{0.2}                                                         & \multicolumn{1}{c|}{0.4}                                                               & 0.4                                                                      \\
\multicolumn{1}{c|}{\textbf{Ms-IV}} & \multicolumn{1}{c|}{\textbf{0.1}}                                                            & \multicolumn{1}{c|}{\textbf{0.1}}                                                              & \multicolumn{1}{c|}{0.2}                                                                  & \multicolumn{1}{c|}{\textbf{0.0}}                                                      & \textbf{0.1}                                                            
\end{tabular}
}
\subcaption{}
\label{tab:localization}
\end{minipage}
~
\begin{minipage}{0.25\textwidth}
\centering
\renewcommand{\arraystretch}{0.65}
\centering
\resizebox{\textwidth}{!}{
\begin{tabular}{c|c|c}
\rowcolor[HTML]{F2DEDD} 
\textbf{Mean VGG}       & \textbf{LIME} & \textbf{Ms-IV}                       \\ \hline
\textbf{Background}     & 0.33          & {\color[HTML]{333333} \textbf{0.25}} \\ \hline
\textbf{Top 2 concepts} & 0.20          & \textbf{0.21}                        \\ \hline
\rowcolor[HTML]{F2DEDD} 
\textbf{Mean ResNet}    & \textbf{LIME} & \textbf{Ms-IV}                       \\ \hline
\textbf{Background}     & 0.39          & {\color[HTML]{333333} \textbf{0.26}} \\ \hline
\textbf{Top 2 concepts} & 0.15          & \textbf{0.19}                       
\end{tabular}
}
\subcaption{}
\label{tab:cub_dataset_localization_summary}
\end{minipage}

\begin{minipage}{0.9\textwidth}
\centering
\renewcommand{\arraystretch}{0.5}
\centering
\resizebox{.8 \textwidth}{!}{
\begin{tabular}{c|c|lr|lr|c|c|lr|cc}
\rowcolor[HTML]{DAE8FC} 
\textbf{Cluster}             & \textbf{Method} & \multicolumn{2}{c|}{\cellcolor[HTML]{DAE8FC}\textbf{VGG}}                                                                                    & \multicolumn{2}{c|}{\cellcolor[HTML]{DAE8FC}\textbf{ResNet}}                                                                                   & \cellcolor[HTML]{DAE8FC}\textbf{Cluster} & \cellcolor[HTML]{DAE8FC}\textbf{Method} & \multicolumn{2}{c|}{\cellcolor[HTML]{DAE8FC}\textbf{VGG}}                                                                                    & \multicolumn{2}{c}{\cellcolor[HTML]{DAE8FC}\textbf{ResNet}}                                                                                                                           \\ \hline
                             & \textbf{LIME}   & \begin{tabular}[c]{@{}l@{}}Background\\ Nape\\ Back\end{tabular}       & \begin{tabular}[c]{@{}r@{}}0.26\\ 0.11\\ 0.10\end{tabular}          & \begin{tabular}[c]{@{}l@{}}Background\\ Nape\\ Left wing\end{tabular}    & \begin{tabular}[c]{@{}r@{}}0.33\\ 0.11\\ 0.07\end{tabular}          &                                          & \textbf{LIME}                           & \begin{tabular}[c]{@{}l@{}}Background\\ Crown\\ Back\end{tabular}      & \textbf{\begin{tabular}[c]{@{}r@{}}0.24\\ 0.17\\ 0.09\end{tabular}} & \multicolumn{1}{l}{\begin{tabular}[c]{@{}l@{}}Background\\ Left wing\\ Crown\end{tabular}}  & \multicolumn{1}{r}{\begin{tabular}[c]{@{}r@{}}0.40\\ 0.08\\ 0.07\end{tabular}}          \\
\multirow{-2}{*}{\textbf{0}} & \textbf{Ms-IV}  & \begin{tabular}[c]{@{}l@{}}Background\\ Back\\ Crown\end{tabular}      & \textbf{\begin{tabular}[c]{@{}r@{}}0.09\\ 0.13\\ 0.12\end{tabular}} & \begin{tabular}[c]{@{}l@{}}Background\\ Right wing\\ Tail\end{tabular}   & \textbf{\begin{tabular}[c]{@{}r@{}}0.21\\ 0.11\\ 0.09\end{tabular}} & \multirow{-2}{*}{\textbf{6}}             & \textbf{Ms-IV}                          & \begin{tabular}[c]{@{}l@{}}Background\\ Breast\\ Nape\end{tabular}     & \begin{tabular}[c]{@{}r@{}}0.28\\ 0.10\\ 0.09\end{tabular}          & \multicolumn{1}{l}{\begin{tabular}[c]{@{}l@{}}Background\\ Right wing\\ Beak\end{tabular}}  & \multicolumn{1}{r}{\textbf{\begin{tabular}[c]{@{}r@{}}0.26\\ 0.11\\ 0.08\end{tabular}}} \\ \hline
                             & \textbf{LIME}   & \begin{tabular}[c]{@{}l@{}}Background\\ Nape\\ Back\end{tabular}       & \begin{tabular}[c]{@{}r@{}}0.21\\ 0.11\\ 0.09\end{tabular}          & \begin{tabular}[c]{@{}l@{}}Background\\ Tail\\ Left wing\end{tabular}    & \begin{tabular}[c]{@{}r@{}}0.55\\ 0.07\\ 0.06\end{tabular}          &                                          & \textbf{LIME}                           & \begin{tabular}[c]{@{}l@{}}Background\\ Crown\\ Forehead\end{tabular}  & \begin{tabular}[c]{@{}r@{}}0.67\\ 0.08\\ 0.05\end{tabular}          & \multicolumn{1}{l}{\begin{tabular}[c]{@{}l@{}}Background\\ Crown\\ Right wing\end{tabular}} & \multicolumn{1}{r}{\begin{tabular}[c]{@{}r@{}}0.31\\ 0.11\\ 0.08\end{tabular}}          \\
\multirow{-2}{*}{\textbf{1}} & \textbf{Ms-IV}  & \begin{tabular}[c]{@{}l@{}}Background\\ Beak\\ Tail\end{tabular}       & \textbf{\begin{tabular}[c]{@{}r@{}}0.18\\ 0.13\\ 0.10\end{tabular}} & \begin{tabular}[c]{@{}l@{}}Background\\ Crown\\ Tail\end{tabular}        & \textbf{\begin{tabular}[c]{@{}r@{}}0.24\\ 0.13\\ 0.10\end{tabular}} & \multirow{-2}{*}{\textbf{7}}             & \textbf{Ms-IV}                          & \begin{tabular}[c]{@{}l@{}}Background\\ Right wing\\ Tail\end{tabular} & \textbf{\begin{tabular}[c]{@{}r@{}}0.43\\ 0.07\\ 0.07\end{tabular}} & \multicolumn{1}{l}{\begin{tabular}[c]{@{}l@{}}Background\\ Nape\\ Back\end{tabular}}        & \multicolumn{1}{r}{\textbf{\begin{tabular}[c]{@{}r@{}}0.29\\ 0.12\\ 0.09\end{tabular}}} \\ \hline
                             & \textbf{LIME}   & \begin{tabular}[c]{@{}l@{}}Background\\ Left wing\\ Nape\end{tabular}  & \begin{tabular}[c]{@{}r@{}}0.36\\ 0.09\\ 0.08\end{tabular}          & \begin{tabular}[c]{@{}l@{}}Background\\ Back\\ Crown\end{tabular}        & \begin{tabular}[c]{@{}r@{}}0.39\\ 0.08\\ 0.07\end{tabular}          &                                          & \textbf{LIME}                           & \begin{tabular}[c]{@{}l@{}}Background\\ Nape\\ Back\end{tabular}       & \textbf{\begin{tabular}[c]{@{}r@{}}0.28\\ 0.12\\ 0.10\end{tabular}} & \multicolumn{1}{l}{\begin{tabular}[c]{@{}l@{}}Background\\ Tail\\ Back\end{tabular}}        & \multicolumn{1}{r}{\textbf{\begin{tabular}[c]{@{}r@{}}0.33\\ 0.08\\ 0.07\end{tabular}}} \\
\multirow{-2}{*}{\textbf{2}} & \textbf{Ms-IV}  & \begin{tabular}[c]{@{}l@{}}Background\\ Nape\\ Left wing\end{tabular}  & \textbf{\begin{tabular}[c]{@{}r@{}}0.14\\ 0.12\\ 0.11\end{tabular}} & \begin{tabular}[c]{@{}l@{}}Background\\ Right wing\\ Beak\end{tabular}   & \textbf{\begin{tabular}[c]{@{}r@{}}0.19\\ 0.11\\ 0.11\end{tabular}} & \multirow{-2}{*}{\textbf{8}}             & \textbf{Ms-IV}                          & \begin{tabular}[c]{@{}l@{}}Background\\ Nape\\ Crown\end{tabular}      & \begin{tabular}[c]{@{}r@{}}0.37\\ 0.09\\ 0.08\end{tabular}          & \multicolumn{1}{l}{\begin{tabular}[c]{@{}l@{}}Background\\ Tail\\ Crown\end{tabular}}       & \multicolumn{1}{r}{\begin{tabular}[c]{@{}r@{}}0.41\\ 0.08\\ 0.07\end{tabular}}          \\ \hline
                             & \textbf{LIME}   & \begin{tabular}[c]{@{}l@{}}Background\\ Nape\\ Crown\end{tabular}      & \textbf{\begin{tabular}[c]{@{}r@{}}0.15\\ 0.14\\ 0.14\end{tabular}} & \begin{tabular}[c]{@{}l@{}}Background\\ Beak\\ Back\end{tabular}         & \begin{tabular}[c]{@{}r@{}}0.38\\ 0.08\\ 0.08\end{tabular}          &                                          & \textbf{LIME}                           & \begin{tabular}[c]{@{}l@{}}Background\\ Tail\\ Back\end{tabular}       & \textbf{\begin{tabular}[c]{@{}r@{}}0.28\\ 0.10\\ 0.10\end{tabular}} & -                                                                                           & -                                                                                       \\
\multirow{-2}{*}{\textbf{3}} & \textbf{Ms-IV}  & \begin{tabular}[c]{@{}l@{}}Background\\ Right wing\\ Back\end{tabular} & \begin{tabular}[c]{@{}r@{}}0.23\\ 0.10\\ 0.10\end{tabular}          & \begin{tabular}[c]{@{}l@{}}Background\\ Right wing\\ Breast\end{tabular} & \textbf{\begin{tabular}[c]{@{}r@{}}0.26\\ 0.09\\ 0.09\end{tabular}} & \multirow{-2}{*}{\textbf{9}}             & \textbf{Ms-IV}                          & \begin{tabular}[c]{@{}l@{}}Background\\ Back\\ Tail\end{tabular}       & \begin{tabular}[c]{@{}r@{}}0.36\\ 0.09\\ 0.08\end{tabular}          & -                                                                                           & -                                                                                       \\ \hline
                             & \textbf{LIME}   & \begin{tabular}[c]{@{}l@{}}Background\\ Back\\ Crown\end{tabular}      & \begin{tabular}[c]{@{}r@{}}0.32\\ 0.12\\ 0.11\end{tabular}          & \begin{tabular}[c]{@{}l@{}}Background\\ Left wing\\ Back\end{tabular}    & \begin{tabular}[c]{@{}r@{}}0.47\\ 0.06\\ 0.05\end{tabular}          &                                          & \textbf{LIME}                           & \begin{tabular}[c]{@{}l@{}}Background\\ Back\\ Right wing\end{tabular} & \begin{tabular}[c]{@{}r@{}}0.64\\ 0.09\\ 0.05\end{tabular}          & -                                                                                           & -                                                                                       \\
\multirow{-2}{*}{\textbf{4}} & \textbf{Ms-IV}  & \begin{tabular}[c]{@{}l@{}}Background\\ Right wing\\ Beak\end{tabular} & \textbf{\begin{tabular}[c]{@{}r@{}}0.25\\ 0.12\\ 0.09\end{tabular}} & \begin{tabular}[c]{@{}l@{}}Background\\ Beak\\ Nape\end{tabular}         & \textbf{\begin{tabular}[c]{@{}r@{}}0.25\\ 0.12\\ 0.10\end{tabular}} & \multirow{-2}{*}{\textbf{10}}            & \textbf{Ms-IV}                          & \begin{tabular}[c]{@{}l@{}}Background\\ Right wing\\ Tail\end{tabular} & \textbf{\begin{tabular}[c]{@{}r@{}}0.30\\ 0.13\\ 0.09\end{tabular}} & \textbf{-}                                                                                  & -                                                                                       \\ \hline
                             & \textbf{LIME}   & \begin{tabular}[c]{@{}l@{}}Background\\ Back\\ Crown\end{tabular}      & \begin{tabular}[c]{@{}r@{}}0.42\\ 0.10\\ 0.09\end{tabular}          & \begin{tabular}[c]{@{}l@{}}Background\\ Nape\\ Beak\end{tabular}         & \begin{tabular}[c]{@{}r@{}}0.40\\ 0.08\\ 0.07\end{tabular}          &                                          & \textbf{LIME}                           & \begin{tabular}[c]{@{}l@{}}Background\\ Left wing\\ Tail\end{tabular}  & \begin{tabular}[c]{@{}r@{}}0.21\\ 0.11\\ 0.10\end{tabular}          & -                                                                                           & -                                                                                       \\
\multirow{-2}{*}{\textbf{5}} & \textbf{Ms-IV}  & \begin{tabular}[c]{@{}l@{}}Background\\ Right wing\\ Nape\end{tabular} & \textbf{\begin{tabular}[c]{@{}r@{}}0.25\\ 0.13\\ 0.12\end{tabular}} & \begin{tabular}[c]{@{}l@{}}Background\\ Right wing\\ Back\end{tabular}   & \textbf{\begin{tabular}[c]{@{}r@{}}0.25\\ 0.09\\ 0.09\end{tabular}} & \multirow{-2}{*}{\textbf{11}}            & \textbf{Ms-IV}                          & \begin{tabular}[c]{@{}l@{}}Background\\ Right wing\\ Tail\end{tabular} & \textbf{\begin{tabular}[c]{@{}r@{}}0.22\\ 0.17\\ 0.11\end{tabular}} & \textbf{-}                                                                                  & -                                                                                      
\end{tabular}
}
\subcaption{}
\label{tab:cub_dataset_localization}
\end{minipage}
\end{table}

These results evince better localization using Ms-IV with fewer background highlights. We also present the top 2 most frequent concepts in each cluster. The ideal is to have a high percentage of images highlighting a few parts. Ms-IV presents an improvement in LIME results, especially for ResNet model.

\subsection{Knowledge Discovery}

\begin{figure}[!ht]
  \centering
  \begin{tabular}{c}
  \begin{minipage}{0.4\textwidth}
  \includegraphics[width=\linewidth]{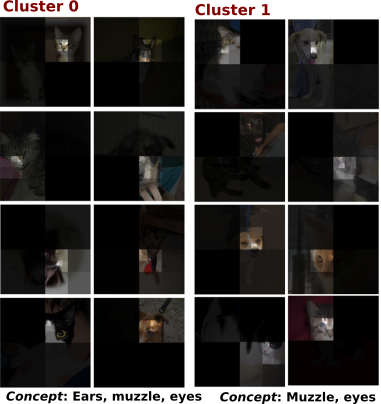}
\subcaption{Visualizations 2 clusters ResNet-18}
\end{minipage}
~
\begin{minipage}{0.5\textwidth}
  \includegraphics[width=\linewidth]{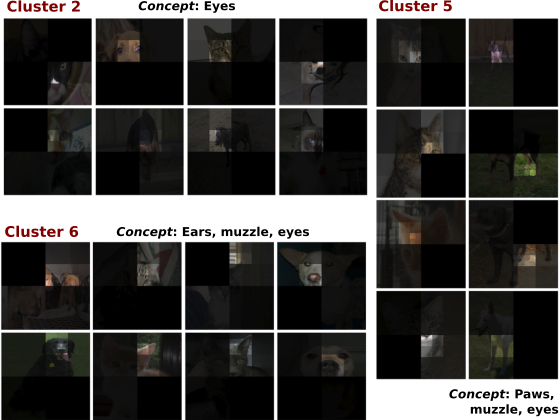}
\subcaption{Visualizations 3 clusters VGG16}
\end{minipage}
\end{tabular}
  \caption{Some visualizations obtained for clusters 0 and 1 of ResNet-18, and clusters 2, 5, and 6 of VGG16 (other visualizations in~\ref{ap:knowledge}). We present the selected concepts for these clusters, by 24 participants, to describe the two classes. According to the answers, for ResNet-18: cluster 0 presents the \textbf{eye} and \textbf{muzzle} of cats, highlighting \textbf{eye} and \textbf{ear} of dogs. Cluster 1 presents \textbf{eye} for both classes and \textbf{muzzle} for dogs. For VGG16: cluster 2  presents \textbf{eye} for both classes. Cluster 5 detects the \textbf{eye} for cats, and the \textbf{muzzle} and \textbf{paws} for dogs. Cluster 6 presents \textbf{ears} of cats and, the \textbf{muzzle} and \textbf{eyes} for dogs.}
  \label{fig:kd_cluster01_main}
\end{figure}

In these experiments, we \revFour{applied} the ensemble of methods to \textit{find concepts} and \textit{detect bias}. To measure the provided interpretability, the produced visualizations were analyzed by the 24 individuals selected from computer and non-computer experts from two countries: Brazil and France. There were a total of 11 computer experts, including people from industry and academia. There were a total of 13 non-computer experts including people from non-academic domains and from the three main domains (in similar quantities): humanities (sociology, geography, and architecture), biological sciences (medicine, physical education), and exact sciences (mathematics) in different educational stages (undergraduate, graduate, and post-graduate).

\textbf{Finding concepts:} We selected six MAG-generated clusters from ResNet-18 and VGG16. We visualized each cluster through Ms-IV applied to 16 images (8 cats and 8 dogs) from the top-middle-ranked positions. From a ranking of 512 images, we started at position 100 to avoid sparsity in higher and lower positions (possible outliers). We presented the Ms-IV visualizations of these image subsets to the research participants and asked which animal part corresponded to the lighter regions in dogs and cats. There were a total of 12 image subsets (limited analysis to six clusters per network).

Of the 13 concepts, fewer than three of them received most of the participants' votes per cluster. There was an agreement about concepts for both computer and non-computer experts. Concepts such as \textbf{eyes} and \textbf{muzzle} were the most frequently observed. We highlight Fig.~\ref{fig:kd_cluster01_main} as an example of high agreement and variability of concepts: \textbf{eye}, \textbf{muzzle}, \textbf{paws}, and \textbf{ear}.%

\begin{table}[!ht]
\caption{From a total of 24 participants and 8 different bias/non-bias comparisons, $77\%$ of the responses showed the non-bias group choice as a better separation (results using the cat vs. dog model). We display the values for computer (\textbf{Comp.}) and non-computer (\textbf{Non-comp.}) experts to make a selection between \textbf{Bias} and \textbf{Non-bias}. Even non-computer experts present a high percentage of the non-bias choice.}
\label{tab:bias_analysis}
\centering
\renewcommand{\arraystretch}{0.65}
\resizebox{.65 \textwidth}{!}{
\begin{tabular}{crrrcrrr}
\multicolumn{1}{l}{}                                             & \multicolumn{1}{c}{\textbf{Comp.}} & \multicolumn{1}{c}{\textbf{Non-Comp.}} & \multicolumn{1}{c}{\textbf{Total}}              & \textbf{}                                                            & \multicolumn{1}{c}{\textbf{Comp.}} & \multicolumn{1}{c}{\textbf{Non-Comp.}} & \multicolumn{1}{c}{\textbf{Total}} \\ \cline{2-8} 
\multicolumn{1}{c|}{\textbf{Bias 0}}                             & 3                                  & 4                                      & \multicolumn{1}{r|}{7}                          & \multicolumn{1}{c|}{\textbf{Bias 4}}                                 & 0                                  & 1                                      & 1                                  \\
\rowcolor[HTML]{DAE8FC} 
\multicolumn{1}{c|}{\cellcolor[HTML]{DAE8FC}\textbf{Non-bias 0}} & 8                                  & 9                                      & \multicolumn{1}{r|}{\cellcolor[HTML]{DAE8FC}17} & \multicolumn{1}{c|}{\cellcolor[HTML]{DAE8FC}\textbf{Non-bias 4}}     & 11                                 & 12                                     & 23                                 \\
\multicolumn{1}{c|}{\textbf{Bias 1}}                             & 3                                  & 3                                      & \multicolumn{1}{r|}{6}                          & \multicolumn{1}{c|}{\textbf{Bias 5}}                                 & 2                                  & 4                                      & 6                                  \\
\rowcolor[HTML]{DAE8FC} 
\multicolumn{1}{c|}{\cellcolor[HTML]{DAE8FC}\textbf{Non-bias 1}} & 8                                  & 10                                     & \multicolumn{1}{r|}{\cellcolor[HTML]{DAE8FC}18} & \multicolumn{1}{c|}{\cellcolor[HTML]{DAE8FC}\textbf{Non-bias 5}}     & 9                                  & 9                                      & 18                                 \\
\multicolumn{1}{c|}{\textbf{Bias 2}}                             & 2                                  & 4                                      & \multicolumn{1}{r|}{6}                          & \multicolumn{1}{c|}{\textbf{Bias 6}}                                 & 2                                  & 5                                      & 7                                  \\
\rowcolor[HTML]{DAE8FC} 
\multicolumn{1}{c|}{\cellcolor[HTML]{DAE8FC}\textbf{Non-bias 2}} & 9                                  & 9                                      & \multicolumn{1}{r|}{\cellcolor[HTML]{DAE8FC}18} & \multicolumn{1}{c|}{\cellcolor[HTML]{DAE8FC}\textbf{Non-bias 6}}     & 9                                  & 8                                      & 17                                 \\
\multicolumn{1}{c|}{\textbf{Bias 3}}                             & 0                                  & 0                                      & \multicolumn{1}{r|}{0}                          & \multicolumn{1}{c|}{\textbf{Bias 7}}                                 & 3                                  & 7                                      & 10                                 \\
\rowcolor[HTML]{DAE8FC} 
\multicolumn{1}{c|}{\cellcolor[HTML]{DAE8FC}\textbf{Non-bias 3}} & 11                                 & 13                                     & \multicolumn{1}{r|}{\cellcolor[HTML]{DAE8FC}24} & \multicolumn{1}{c|}{\cellcolor[HTML]{DAE8FC}\textbf{Non-bias 7}}     & 8                                  & 6                                      & 14                                 \\ \cline{2-8} 
\multicolumn{1}{l}{}                                             & \multicolumn{1}{l}{}               & \multicolumn{1}{l}{}                   & \multicolumn{1}{l}{}                            & \multicolumn{1}{c|}{\textbf{Bias Total}}                             & 15                                 & 28                                     & 43                                 \\
\multicolumn{1}{l}{}                                             & \multicolumn{1}{l}{}               & \multicolumn{1}{l}{}                   & \multicolumn{1}{l}{}                            & \multicolumn{1}{c|}{\cellcolor[HTML]{DAE8FC}\textbf{Non-bias Total}} & \cellcolor[HTML]{DAE8FC}73         & \cellcolor[HTML]{DAE8FC}76             & \cellcolor[HTML]{DAE8FC}149        \\ \cline{5-8} 
\multicolumn{1}{l}{}                                             & \multicolumn{1}{l}{}               & \multicolumn{1}{l}{}                   & \multicolumn{1}{l}{}                            & \multicolumn{1}{c|}{\textit{\textbf{\% Non-bias Total}}}             & \cellcolor[HTML]{FFFFC7}82\%       & \cellcolor[HTML]{FFFFC7}73\%           & \cellcolor[HTML]{FFFFC7}77\%      
\end{tabular}
}
\end{table}

\textbf{Bias Detection:} We \revFour{compared} a biased and a less biased model (more accurate). The analyzed ResNet-18 model \revFour{was} the less biased, we call it the normal one. We trained an extra ResNet-18 model, initialized with ImageNet weights, and trained with 100 images, 50 dark cats, and 50 beige dogs.

We generated the biased and unbiased ResNet-18 image subsets to each concept (as in the \textbf{finding concepts} part). We paired one biased ResNet-18 group and one normal ResNet-18 group. We asked the participants which of the models seemed to highlight only the important parts to differentiate cats and dogs, without explaining Neural Network bias. 

Results in Table~\ref{tab:bias_analysis} show, that both computer and non-computer experts found, with high accuracy, the unbiased model  ($73\%$ of correct for the non-computer experts). Our visualization facilitates model analysis, even for non-specialists.
\section{Conclusion}
\label{sec:conclusion}

We propose a \revThree{novel} way to make CNNs interpretable thanks to the combination of \emph{MAGE}, to group feature maps into \textquote{concepts}, and \emph{Ms-IV}, to provide a simple (multiscale) understanding of \revThree{a given} model's knowledge. \revThree{The} \revFour{$\CAOC$} metric \revThree{considers} the structure and organization of the model's final decision space and provides global awareness of sample perturbations. In the future, we plan to improve the \textquote{\revTwo{clusters} quality} through hierarchical/spectral clustering techniques, and to introduce more subtle segmentation techniques using mathematical morphology in our multiscale visualization.

\section*{Acknowledgements}

We extend our gratitude to Professor Didier Verna, Dr. Samuel Feng, and Sarah Almeida Carneiro for generously dedicating time to review our article and for engaging in insightful discussions with us.

\bibliographystyle{elsarticle-num} 
\bibliography{main}

\begin{thebibliography}{10}
\expandafter\ifx\csname url\endcsname\relax
  \def\url#1{\texttt{#1}}\fi
\expandafter\ifx\csname urlprefix\endcsname\relax\def\urlprefix{URL }\fi
\expandafter\ifx\csname href\endcsname\relax
  \def\href#1#2{#2} \def\path#1{#1}\fi

\bibitem{Abeyagunasekera:2022:i2ct}
S.~H.~P. Abeyagunasekera, Y.~Perera, K.~Chamara, U.~Kaushalya, P.~Sumathipala,
  O.~Senaweera, Lisa : Enhance the explainability of medical images unifying
  current xai techniques, in: 7th International conference for Convergence in
  Technology (I2CT), 2022, pp. 1--9.

\bibitem{adadi:IEEEAccess:2018}
A.~Adadi, M.~Berrada, Peeking inside the black-box: A survey on explainable
  artificial intelligence {(xAI)}, IEEE Access 6 (2018) 1--23.

\bibitem{ali:2023:infoFusion}
S.~Ali, T.~Abuhmed, S.~El-Sappagh, K.~Muhammad, J.~M. Alonso-Moral,
  R.~Confalonieri, R.~Guidotti, J.~{Del Ser}, N.~Díaz-Rodríguez, F.~Herrera,
  Explainable artificial intelligence ({XAI}): What we know and what is left to
  attain trustworthy artificial intelligence, Information Fusion 99 (2023)
  101805.

\bibitem{aslam:2022:interpretable}
N.~Aslam, I.~U. Khan, S.~Mirza, A.~AlOwayed, F.~M. Anis, R.~M. Aljuaid,
  R.~Baageel, Interpretable machine learning models for malicious domains
  detection using explainable artificial intelligence (xai), Sustainability
  14~(12) (2022) 7375.

\bibitem{bach:plosOne:2015}
S.~Bach, A.~Binder, G.~Montavon, F.~Klauschen, K.-R. Muller, W.~Samek, On
  pixel-wise explanations for non-linear classifier decisions by layer-wise
  relevance propagation, PLoS One 10~(7) (2015) 1--46.

\bibitem{bau:2017:cvpr}
D.~Bau, B.~Zhou, A.~Khosla, A.~Oliva, A.~Torralba, Network dissection:
  Quantifying interpretability of deep visual representations, in: 2017 IEEE
  Conference on Computer Vision and Pattern Recognition (CVPR), 2017, pp.
  3319--3327.

\bibitem{bommer:2023:corr}
P.~Bommer, M.~Kretschmer, A.~Hedstr{\"{o}}m, D.~Bareeva, M.~M. H{\"{o}}hne,
  Finding the right {XAI} method - {A} guide for the evaluation and ranking of
  explainable {AI} methods in climate science, CoRR 2303.00652 (2023).

\bibitem{born:2021:ASci}
J.~Born, N.~Wiedemann, M.~Cossio, C.~Buhre, G.~Brändle, K.~Leidermann,
  J.~Goulet, A.~Aujayeb, M.~Moor, B.~Rieck, K.~Borgwardt, Accelerating
  detection of lung pathologies with explainable ultrasound image analysis,
  Applied Sciences 11~(2) (2021).

\bibitem{borys:2023:euRad}
K.~Borys, Y.~A. Schmitt, M.~Nauta, C.~Seifert, N.~Krämer, C.~M. Friedrich,
  F.~Nensa, Explainable ai in medical imaging: An overview for clinical
  practitioners - beyond saliency-based xai approaches, European Journal of
  Radiology (2023).

\bibitem{chaddad:2023:sensors}
A.~Chaddad, J.~Peng, J.~Xu, A.~Bouridane, Survey of explainable ai techniques
  in healthcare, Sensors 23~(2) (2023).

\bibitem{chen:2023:springer}
T.~Chen, Applications of xai for forecasting in the manufacturing domain, in:
  SpringerBriefs in Applied Sciences and Technology, SpringerBriefs in Applied
  Sciences and Technology, Springer Science and Business Media Deutschland
  GmbH, 2023, pp. 13--50.

\bibitem{crook:2023:arxiv}
B.~Crook, M.~Schlüter, T.~Speith, Revisiting the performance-explainability
  trade-off in explainable artificial intelligence (xai), arXiv (2023).

\bibitem{ghorbani:2019:neurips}
A.~Ghorbani, J.~Wexler, J.~Z. Y, B.~Kim, Towards automatic concept-based
  explanations, in: Advances in Neural Information Processing Systems, Vol.~32,
  2019, pp. 1--10.

\bibitem{gu:ieee_med:2021}
R.~Gu, G.~Wang, T.~Song, R.~Huang, M.~Aertsen, J.~Deprest, S.~Ourselin,
  T.~Vercauteren, S.~Zhang, Ca-net: Comprehensive attention convolutional
  neural networks for explainable medical image segmentation, IEEE Transactions
  on Medical Imaging 40 (2021) 699--711.

\bibitem{guidotti:acm:2018}
R.~Guidotti, A.~Monreale, S.~Ruggieri, F.~Turini, F.~Giannotti, D.~Pedreschi, A
  survey of methods for explaining black box models, ACM Computing Surveys
  51~(5) (2018) 1--42.

\bibitem{guyon:2002:machineLearning}
I.~Guyon, J.~Weston, S.~Barnhill, V.~Vapnik, Gene selection for cancer
  classification using support vector machines, Machine Learning 46 (2002)
  389–422.

\bibitem{haghanifar:2022:multimedia}
A.~Haghanifar, M.~M. Majdabadi, Y.~Choi, S.~Deivalakshmi, S.~Ko, Covid-cxnet:
  Detecting covid-19 in frontal chest x-ray images using deep learning,
  Multimedia Tools and Applications 81 (2022) 30615–30645.

\bibitem{He:cvpr:2016}
K.~He, X.~Zhang, S.~Ren, J.~Sun, Deep residual learning for image recognition,
  in: 29th IEEE Conference on Computer Vision and Pattern Recognition (CVPR),
  2016, pp. 770--778.

\bibitem{he:2019:icsvt}
X.~He, Y.~Peng, Fine-grained visual-textual representation learning, IEEE
  Transactions on Circuits and Systems for Video Technology PP (2019) 1--12.

\bibitem{huang:2023:iccv}
W.~Huang, X.~Zhao, G.~Jin, X.~Huang, Safari: Versatile and efficient
  evaluations for robustness of interpretability, in: International Conference
  on Computer Vision (ICCV), 2022, pp. 1--10.

\bibitem{hulsen:2023:ai}
T.~Hulsen, Explainable artificial intelligence (xai): Concepts and challenges
  in healthcare, AI 4~(3) (2023) 652--666.

\bibitem{kim:icml:2018}
B.~Kim, M.~Wattenberg, J.~Gilmer, C.~J. Cai, J.~Wexler, F.~B. Vi{\'e}gas,
  R.~Sayres, Interpretability beyond feature attribution: Quantitative testing
  with {Concept Activation Vectors (TCAV)}, in: 35th International Conference
  on Machine Learning (ICML), 2018, pp. 2668--2677.

\bibitem{li:2018:patternnet}
H.~Li, J.~G. Ellis, L.~Zhang, S.-F. Chang, Patternnet: Visual pattern mining
  with deep neural network, in: Proceedings of the 2018 ACM on international
  conference on multimedia retrieval, 2018, pp. 291--299.

\bibitem{li:2023:ai}
X.~Li, H.~Xiong, X.~Li, X.~Zhang, J.~Liu, H.~Jiang, Z.~Chen, D.~Dou, G-lime:
  Statistical learning for local interpretations of deep neural networks using
  global priors, Artificial Intelligence 314~(C) (2023).

\bibitem{lu:2022:intsystems}
S.~Lu, Z.~Zhu, J.~M. Gorriz, S.-H. Wang, Y.-D. Zhang, Nagnn: Classification of
  covid-19 based on neighboring aware representation from deep graph neural
  network, International Journal of Intelligent Systems 37 (2022) 1572–159.

\bibitem{lundberg:neurips:2017}
S.~M. Lundberg, S.-I. Lee, A unified approach to interpreting model
  predictions, in: 31st International Conference on Neural Information
  Processing Systems (NeurIPS), 2017, pp. 4768--4777.

\bibitem{mcinnes:2018:umap}
L.~McInnes, J.~Healy, J.~Melville, Umap: Uniform manifold approximation and
  projection for dimension reduction, arXiv (2018).

\bibitem{park:cvpr:2018}
D.~H. Park, L.~A. Hendricks, Z.~Akata, A.~Rohrbach, B.~Schiele, T.~Darrell,
  M.~Rohrbach, Multimodal explanations: Justifying decisions and pointing to
  the evidence, in: 31st International Conference on Computer Vision and
  Pattern Recognition (CVPR), 2018, pp. 8779--8788.

\bibitem{quinlan:ml:1986}
J.~R. Quinlan, Induction of decision trees, Machine Learning 1 (1986) 81--106.

\bibitem{Ribeiro:CKDDM:2016}
M.~T. Ribeiro, S.~Singh, C.~Guestrin, "{Why Should I Trust You?}": Explaining
  the predictions of any classifier, in: 22nd International Conference on
  Knowledge Discovery and Data Mining (KDD), 2016, pp. 1135--1144.

\bibitem{priya:2023:icaccs}
P.~D. S, R.~K. K, V.~S, N.~K, A.~K, An overview of interpretability techniques
  for explainable artificial intelligence (xai) in deep learning-based medical
  image analysis, in: 9th International Conference on Advanced Computing and
  Communication Systems (ICACCS), Vol.~1, 2023, pp. 175--182.

\bibitem{schwalbe:2023:dataMining}
G.~Schwalbe, B.~Finzel, A comprehensive taxonomy for explainable artificial
  intelligence: a systematic survey of surveys on methods and concepts, Data
  Mining and Knowledge Discovery (2023).

\bibitem{Selvaraju:ICCV:2017}
R.~R. {Selvaraju}, M.~{Cogswell}, A.~{Das}, R.~{Vedantam}, D.~{Parikh},
  D.~{Batra}, Grad-cam: Visual explanations from deep networks via
  gradient-based localization, in: 16th International Conference on Computer
  Vision (ICCV), 2017, pp. 618--626.

\bibitem{shannon:1948:bell}
C.~E. Shannon, A mathematical theory of communication, The Bell System
  Technical Journal 27 (1948) 379--423.

\bibitem{shen:2021:medical}
Y.~Shen, N.~Wu, J.~Phang, J.~Park, K.~Liu, S.~Tyagi, L.~Heacock, S.~G. Kim,
  L.~Moy, K.~Cho, K.~J. Geras, An interpretable classifier for high-resolution
  breast cancer screening images utilizing weakly supervised localization,
  Medical Image Analysis 68 (2021) 101908.

\bibitem{Sheu:2023:access}
R.-K. Sheu, M.~Pardeshi, K.-C. Pai, L.-C. Chen, C.-L. Wu, W.-C. Chen,
  Interpretable classification of pneumonia infection using explainable ai
  (xai-icp), IEEE Access PP (2023) 1--1.

\bibitem{shrikumar:icml:2017}
A.~Shrikumar, P.~Greenside, A.~Kundaje, Learning important features through
  propagating activation differences, in: 34th International Conference on
  Machine Learning (ICML), 2017, pp. 3145--3153.

\bibitem{Simonyan:iclr:2015}
K.~Simonyan, A.~Zisserman, Very deep convolutional networks for large-scale
  image recognition, in: 3rd International Conference on Learning
  Representations (ICLR), 2015, pp. 1--14.

\bibitem{Springenberg:ICLR:2015}
J.~T. Springenberg, A.~Dosovitskiy, T.~Brox, M.~Riedmiller, Striving for
  simplicity: The all convolutional net, in: 3rd International Conference on
  Learning Representations (ICLR), 2015, pp. 1--14.

\bibitem{Sundararajan:2017:icml}
M.~Sundararajan, A.~Taly, Q.~Yan, Axiomatic attribution for deep networks, in:
  34th International Conference on Machine Learning (ICML), 2017, pp.
  3319--3328.

\bibitem{sutton:2022:sciReport}
R.~T. Sutton, O.~R. Zaiane, R.~Goebel, D.~C. Baumgart, Artificial intelligence
  enabled automated diagnosis and grading of ulcerative colitis endoscopy
  images, Scientific Reports 12~(2748) (2022).

\bibitem{tan:2022:neuralNets}
R.~Tan, L.~Gao, N.~Khan, L.~Guan, Interpretable artificial intelligence through
  locality guided neural networks, Neural Networks 155 (2022) 58--73.

\bibitem{thiagarajan:arxiv:2016}
J.~J. Thiagarajan, B.~Kailkhura, P.~Sattigeri, K.~N. Ramamurthy, Treeview:
  Peeking into deep neural networks via feature-space partitioning (2016).

\bibitem{zeiler:eccv:2014}
M.~D. Zeiler, R.~Fergus, Visualizing and understanding convolutional networks,
  in: 13th European Conference on Computer Vision (ECCV), 2014, pp. 818--833.

\bibitem{zhang:aaai:2018}
Q.~Zhang, R.~Cao, F.~Shi, Y.~N. Wu, S.-C. Zhu, Interpreting cnn knowledge via
  an explanatory graph, in: 32nd AAAI Conference on Artificial Intelligence
  (AAAI), 2018, pp. 4454--4463.

\bibitem{zhang:2021:aaai}
R.~Zhang, P.~Madumal, T.~Miller, K.~A. Ehinger, B.~I.~P. Rubinstein, Invertible
  concept-based explanations for {CNN} models with non-negative concept
  activation vectors, in: 35th Conference on Artificial Intelligence (AAAI),
  2021, pp. 11682 -- 11690.

\bibitem{zhou:cvpr:2016}
B.~Zhou, A.~Khosla, A.~Lapedriza, A.~Oliva, A.~Torralba, Learning deep features
  for discriminative localization, in: 29th Conference on Computer Vision and
  Pattern Recognition (CVPR), 2016, pp. 2921--2929.

\bibitem{Zhou:2021:sigkdd}
Z.~Zhou, G.~Hooker, F.~Wang, S-lime: Stabilized-lime for model explanation, in:
  27th Conference on Knowledge Discovery \& Data Mining (SIGKDD), 2021, pp.
  1--10.

\end{thebibliography}

\newpage
\appendix

\section{Decomposition of the feature space into concepts: Maximum Activation Groups Extraction (MAGE)}
\label{ap:mage}

\begin{figure}[!ht]
  \centering
  \includegraphics[width=0.7\linewidth]{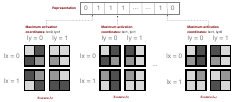}
  \caption{Finding the \textquote{position} of a feature $\featureimagecarres$ (in each image of) the data set (illustrative example). We start from a sequence $(\image_i)_i$ of images whose domain is of size $4 \times 4$. We decompose their common domain into $4$ patches of the same size $2 \times 2$. On the first image, we observe that among the 4 four subdivisions in the $\featureimagecarres^{th}$ feature map, the one that maximizes its $1$-norm corresponds to $(\lx = 1,\ly = 1)$, so we write in the vector (depicted above) these values: $1$ and then $1$. We continue this procedure for the next images until we reach the end of the dataset. This vector represents then where the $\featureimagecarres^{th}$ feature is located in the images of the data set; we call it the \emph{representative} of the feature number $\featureimagecarres$.}
  \label{fig:mag_diagram_sup}
\end{figure}

We present in Figure~\ref{fig:mag_diagram_sup} a schematic example of the proposed representation in the MAGE process.

\section{Global causality-based visualization: Multiscale- Interpretable Visualization (Ms-IV)}
\label{ap:msiv}

\begin{figure}[!ht]
  \centering
  \includegraphics[width=0.7\linewidth]{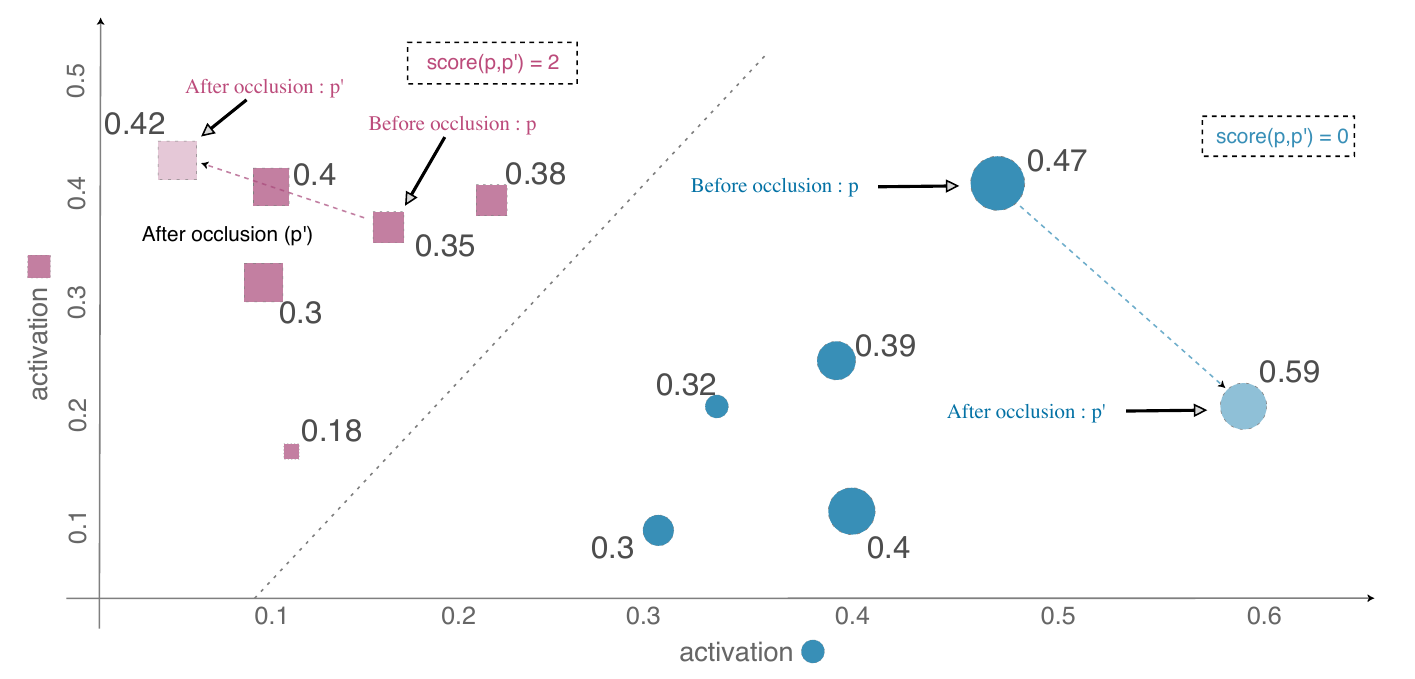}
  \caption{Explanation of the computation of $\CAOC$ (on a fictitious scenario). Here, our dataset is made up of images of one disk or one square, the output is a class (\textquote{disk} or \textquote{square}). We have set the concept value to some random $k$. We plot the activation distributions in a 2D space: the horizontal coordinate represents how much the sample is predicted as being a circle, and vertically, how much it is predicted as being a square. We can see that, by occlusion, one square moves in this space by a distance of $0.07$ and one disk by $0.12$. However, we need some \textit{quantification} of the impact of a concept on the two classes. Thus, we propose using the ranking correlation between before/after the occlusion of the most important patch images relative to the concept $k$. We find that the disk did not move in the disk's ranking when we did the occlusion (it remains the \textquote{strongest} disk), so the correlation $\CAOC(k,\mathit{disks})$ is maximal, meaning that concept $k$ does not have much influence on disks. Conversely, in the case of the square, the square's position changes from $3^{rd}$ to $5^{th}$ position, leading to a ranking change of $2$, thus $\CAOC(k,\mathit{squares})$ is lower, which means that concept $k$ is important for squares.}
  \label{fig:spatial_structure}
\end{figure}

In Figure~\ref{fig:spatial_structure} we visually exemplify the impact of $\CAOC$ in a sparse decision space.

In the sequel, we explain in more detail the algorithm and pseudocode of Ms-IV.

\begin{figure}[!ht]
  \centering
  \includegraphics[width=0.5\linewidth]{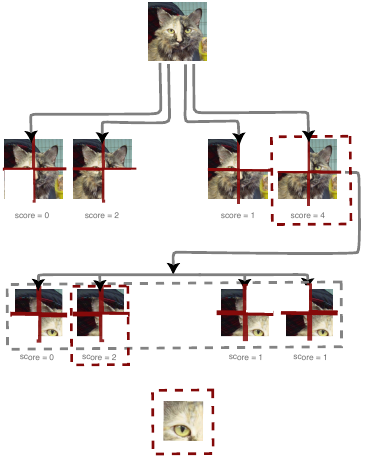}
  \caption{How our visualization algorithm works. We fix some concept value $k$ and some image number $i$. We sort the activations of each image, and we call $\initialposition$ the position of the activation of the current image $\image_i$ in the computed sequence. The goal is to enlighten the areas of the image as much as the concept $k$ is important in each region of this same image. To this aim, we decompose the initial domain into four patches; it is the first step of our recursive subdivision. By occluding separately each of these four patches and computing how much their new position $\newposocc_{\lx,\ly}$ (in the sequence of activations of the occluded images) differs from $\initialposition$, we obtain four scores $\left|\initialposition - \newposocc_{\lx,\ly}\right|$ (called the importance). Choosing the maximal score, allows us to deduce in which of these four patches the concept is represented most. By continuing this recursive subdivision in the most important patches until we reach the minimal size of a patch, we will know how much we have to illuminate each coordinate in the image (by adding up the importances we have computed for each pixel).
  }
  \label{fig:viz_supp}
\end{figure}

We propose here an algorithm (see Algorithm~\ref{alg:ms_iv}  depicted in Figure~\ref{fig:viz_supp}) that uses a multiscale hierarchical approach (such as a quad-tree) capable of highlighting the areas of a given image $\image$ (belonging to $\dataset$) that are important to the network's decision regarding the concept $k$. This approach is \emph{global} in the sense that computations on the entire dataset will have been completed beforehand. Note that our procedure is different from LIME: even though in both cases, we use parts of images to show the model's knowledge, in our case we illuminate the image relative to one unique concept at a time.

For the sake of simplicity, let us introduce a new term: we define an \emph{occlusion} of the image $I$ of domain $\domain$ on a patch $\patch \subseteq \domain$ as $\occlusion_\patch(\image): \domain \rightarrow \Reals$ such that for any $(x,y) \in \domain$, $\occlusion_\patch(\image)(x,y)$ is equal to $\image(x,y)$ when $(x,y) \not \in \patch$, and $0$ otherwise. Now let us formally explain the main steps of our algorithm (\revTwo{we cope} with the recursive part of the algorithm using a list that\revTwo{, to simplify, will not be detailed:}

\begin{enumerate}

\item We fix a visualization threshold ratio $\delta \in \; ]0,1]$, a minimal patch size $\minimalsizepatch \in \Neals^*$ (representing the minimal patch size of a concept), a class $c$, the image $\image^c_{\indexlocalimage}$, and a concept $k$.

\item We compute the sequence $\seq = \sort\left(\left(\activation_c(\image^c_{j},\neuralnetwork_k)\right)_{j \in [1,\nbimages(c)]}\right)$ and then the position $\initialposition$ of $\activation_c(\image^c_{\indexlocalimage},\neuralnetwork_k)$ within it. This position represents \textquote{how much} $\image^c_{\indexlocalimage}$ truly belongs to class $c$.

\item We divide the image into four patches of the same size $\sizepatch = \frac{\imagesize}{2}$ resulting in this partition $\{\PATCHGENERIQUE{0}{0}, \PATCHGENERIQUE{0}{1},\PATCHGENERIQUE{1}{0},\PATCHGENERIQUE{1}{1}\}$ (we assume that the image size is a multiple of $2$).

\item For each $(\lx,\ly) \in \{0,1\}^2$:

\begin{enumerate}    

\item We occlude $\image^c_{\indexlocalimage}$ on the patch $\patch(\lx, \ly,\sizepatch)$ resulting in $\occlusion_{\PATCHGENERIQUE{\lx}{\ly}}(\image^c_{\indexlocalimage})$.

\item We calculate:
$$\seq' := \sort\left(
(\activation_c(\occlusion_{\PATCHGENERIQUE{\lx}{\ly}}(\image^c_{j}),\neuralnetwork_k))_{j \in [1,\nbimages(c)]}
\right)$$
and the position $\newposocc_{\lx,\ly}$ of $\activation_c(\occlusion_{\PATCHGENERIQUE{\lx}{\ly}}(\image^c_{\indexlocalimage}),\neuralnetwork_k)$ in it. It will then represent how much $\image^c_{\indexlocalimage}$ is still of class $c$ after occluding the image in the patch domain. If the initial activation is almost preserved despite the occlusion, we will have $\newposocc_{\lx,\ly} \approx \initialposition$.

\item For this reason, we propose to calculate what we call the \emph{importance} of the patch $\PATCHGENERIQUE{\lx}{\ly}$:
$$\importance(\lx, \ly) = \left|\initialposition - \newposocc_{\lx,\ly}\right|$$

\end{enumerate}

\item We compute a threshold $\thr$ based on $\delta$: $\thr = \max{}\left(\{\importance(\lx, \ly)\}_{(\lx,\ly)}\right) \times \delta$

\item We continue the procedure recursively in the patches that satisfy the inequality $\Imp(\lx,\ly) \geq \thr$ while $\sizepatch$ is greater than or equal to $\minimalsizepatch$.

\item During this recursive procedure, each position $(x,y) \in \domain$ may have been treated several times. We deduce the \emph{accumulated importance} of a position $(x,y)$ relative to the image $\image_i$ by summing all the computed importance terms where this position was occluded. The final result is called the \emph{accumulated importance matrix} and we denote it $\MatriceImportance$.

\item We finally multiply the initial image by $\MatriceImportance$ and we plot it. We have highlighted important regions.

\end{enumerate}

\begin{algorithm}[]
\tiny
\DontPrintSemicolon
{
  \KwInput{$\minimalsizepatch$; $\imagesize$; $\delta$; $I$ of class $c$; concept $k$;}
  \KwOutput{$\MatriceImportance$: matrix of importances}
  \KwData{dataset of squared images $\dataset$}

$\seq := \sort\left(\left(\activation_c(\image^c_{j},\neuralnetwork_k)\right)_{j \in [1,\nbimages(c)]]}\right)$

$\initialposition = \Position(\activation_c(\image^c_{\indexlocalimage},\neuralnetwork_k),\seq)$
  
$\DIMLN := \frac{\imagesize}{\minimalsizepatch}$ \tcp*{dimension of final matrix (smallest patches)}
 
$\MatriceImportance := \CreateMatrixOfZeros(\DIMLN,\DIMLN)$ \tcp*{final matrix initialization}

$\levelmax := int(\sqrt{\DIMLN})$ \tcp*{final level}

$\coords := \left\{\left(0,0\right)\right\}$ \tcp*{level 0 has only patch (0,0)}

$\sizepatch = \imagesize$

\tcc{Quadtree-like propagation}
\For{$\level \in [1, \levelmax]$}
        {
        $\sizepatch := \frac{\sizepatch}{2}$ \tcp*{new patch size}
        
        $\DIMLEVEL := 2^{\level}$ \tcp*{side dimension}
        
        $\MatriceImportanceAuxiliaire := \CreateMatrixOfZeros(\DIMLEVEL,\DIMLEVEL)$\;     
        $\MatriceImportanceAuxiliaireDeux := \CreateMatrixOfZeros(\DIMLN,\DIMLN)$
        
        \tcc{analysis of selected patches}    

        \For{$(\lx,\ly) \in \coords$}{
        \tcc{division into 4 patches}    

            \For{$(\la,\lb) \in [2 \lx , 2 \lx +1] \times [2 \ly, 2 \ly + 1]$}    
                { 
                            $\units := \frac{\DIMLN}{\DIMLEVEL}$ \tcp*{number of smallest patches}

                            $\seq' := \sort\left(
                                (\activation_c(\occlusion_{\PATCHGENERIQUE{\la}{\lb}}(\image^c_{j}),\neuralnetwork_k))_{j \in [1,\nbimages(c)]}
                                \right)$
                            
                            $\newposocc_{\la,\lb} := \Position(\activation_c(\occlusion_{\PATCHGENERIQUE{\la}{\lb}}(\image^c_{\indexlocalimage}),\neuralnetwork_k),\seq')$

                            $\Imp = |\initialposition - \newposocc_{\la,\lb}|$ \tcp*{patch importance}
                               
                           $\MatriceImportanceAuxiliaire(\la;\ \lb) \pluseq \Imp$

                           $\MatriceImportanceAuxiliaireDeux(\la \units,\dots,(\la+1) \units - 1;\ \lb \units,\dots,(\lb+1) \units - 1) \pluseq  \Imp$
                        
                }
        }
        $\MatriceImportanceAuxiliaire := \frac{\MatriceImportanceAuxiliaire - \min{\MatriceImportanceAuxiliaire}}{\max{\MatriceImportanceAuxiliaire} - \min{\MatriceImportanceAuxiliaire}}$ \tcp*{normalization}
        
        $\MatriceImportance \pluseq \MatriceImportanceAuxiliaire$ 

         \tcc{choice of patches for next level}
         
        $\coordsaux := \{\}$

        $\thr = \max(\MatriceImportanceAuxiliaire) \times \delta$ \tcp*{finding threshold value for selection}

        \For{$(\la,\lb) \in [1,\DIMLEVEL] \times [1,\DIMLEVEL]$}{
            \If{$\MatriceImportanceAuxiliaire(\la;\ \lb) \geq \thr$}{
                $\coordsaux := \coordsaux \cup \{(\la, \lb)\}$
        }}

        $\coords := \coordsaux$
        
        \If{$\thr = 0 $}{
        \textbf{break} \tcp*{if no changes in ranking, we early stop}
        }
}
}
\caption{Ms-IV algorithm}
\label{alg:ms_iv}
\end{algorithm}

\newpage
\section{Methods evaluation}
\label{ap:eval}

We present the experiments to test MAGE, $\CAOC$, and Ms-IV, using two CNN architectures, ResNet-18~\cite{He:cvpr:2016} and VGG16~\cite{Simonyan:iclr:2015}, trained on a classification dataset of cats vs. dogs~\footnote{https://www.kaggle.com/competitions/dogs-vs-cats-redux-kernels-edition/data}.

\subsection{Dataset and training}
\label{ap:data_training}

\begin{table}[!ht]
\caption{Accuracy values in training and validation sets for ResNet-18 and VGG16. We present the results for each class separately and together (total accuracy). VGG16 \revFour{presents} the best accuracy.}
\label{tab:acc}
\centering
\begin{tabular}{ccrr|crr}
                                     & \multicolumn{3}{c|}{\textbf{Train}}                                                                    & \multicolumn{3}{c}{\textbf{Val}}                                                                      \\ \cline{2-7} 
\multicolumn{1}{c|}{}                & \textbf{Cat}               & \multicolumn{1}{c}{\textbf{Dog}} & \multicolumn{1}{c|}{\textbf{Total}} & \textbf{Cat}               & \multicolumn{1}{c}{\textbf{Dog}} & \multicolumn{1}{c}{\textbf{Total}} \\ \hline
\multicolumn{1}{c|}{\textbf{ResNet}} & \multicolumn{1}{r}{98.60} & 97.82                            & 98.21                             & \multicolumn{1}{r}{97.93} & 97.79                            & 97.86                            \\
\multicolumn{1}{c|}{\textbf{VGG}}    & \multicolumn{1}{r}{99.09} & 99.00                            & 99.04                             & \multicolumn{1}{r}{98.47} & 98.74                           & 98.61                           
\end{tabular}
\end{table}

We used 19,891 (9,936 dogs and 9,955 cats) images in the training set, 5,109 (2,564 dogs and 2,545 cats) images in the validation set, and 12,499 in the test set. For the training, we used the pre-trained networks in the ImageNet dataset. We excluded the networks' original classification layer and included a 2-neuron layer followed by softmax activation. The networks were trained using Cross Entropy loss, Adam optimizer, and a learning rate of $1e-7$. We saved only the model that minimizes validation loss. We present in Table~\ref{tab:acc} the accuracy values for each model.

\newpage

\subsection{Evaluating the quality of MAGE}
\label{ap:mag_exp}

\begin{figure}[]
\centering
\begin{tabular}{cc}
\includegraphics[width=6.7cm]{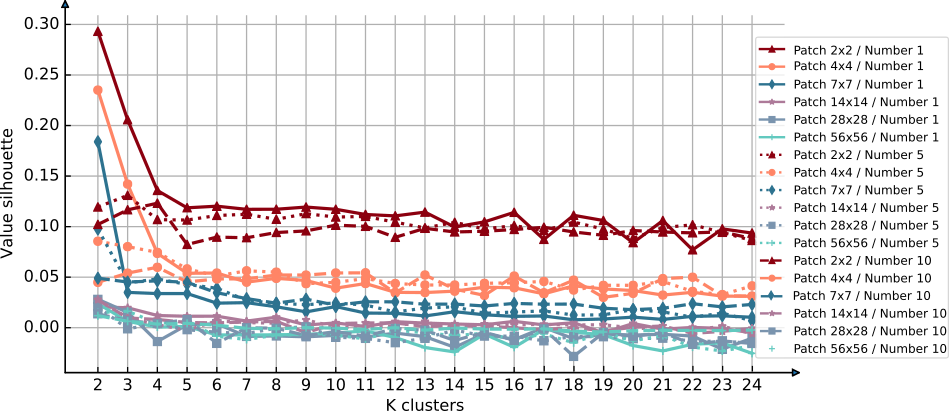}&
\includegraphics[width=6.0cm]{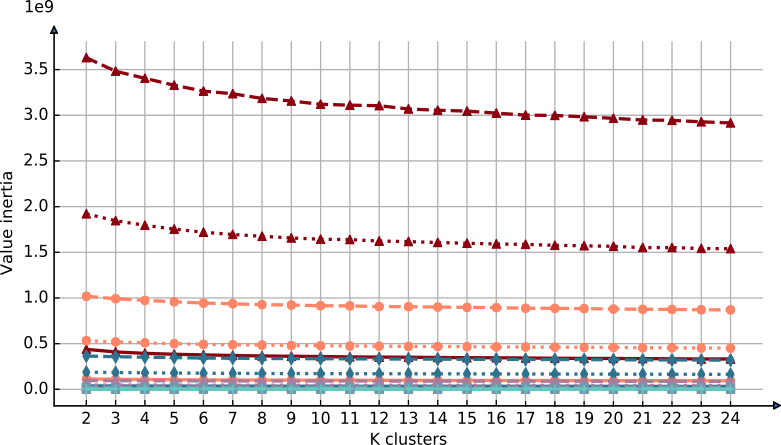}\\
(a) VGG -- Silhouette & (b) VGG -- Inertia\\
\includegraphics[width=6.0cm]{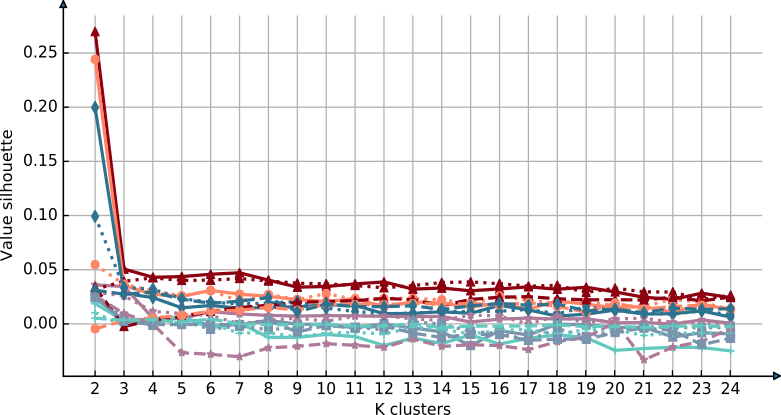}&
\includegraphics[width=6.0cm]{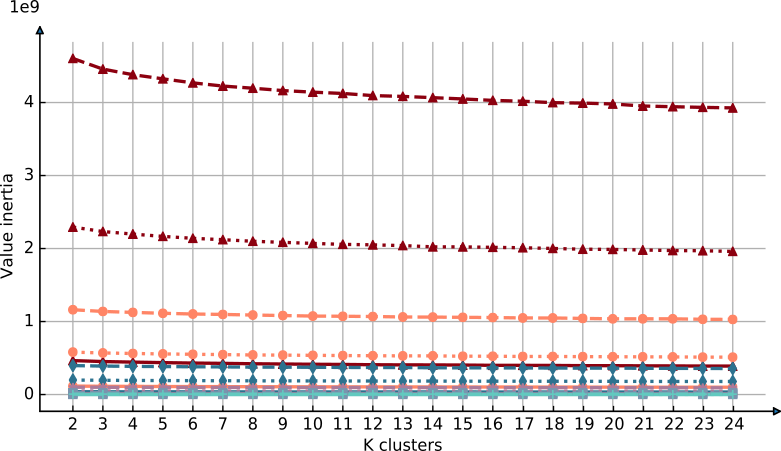}\\
(c) ResNet -- Silhouette & (d) ResNet -- Inertia
\end{tabular}
\caption{Silhouette and Inertia results from K-means with $k \in \left[2,25\right[$ with different sizes and numbers of patches used in the feature map representation, $n \in \{2,4,7,14,28,56\}$, and $t \in \{1,5,10\}$. Figures (a) and (c) present the Silhouette for VGG16 and ResNet respectively. Figures (b) and (d) present the Inertia for VGG16 and ResNet respectively. The best clusters should maximize the Silhouette, which is between -1 and 1. Inertia is not directly comparable, as we changed representation, but will be used for finding the best number of clusters $k$. The representations using small patch sizes seem to improve this mentioned quality.
}
\label{fig:silhouette_inertia}
\end{figure}

\begin{figure}[]
\centering
\begin{tabular}{cc}
\includegraphics[width=6.0cm]{scatter4_512images_no_bias_vgg_no_bias.png} & 
\includegraphics[width=6.0cm]{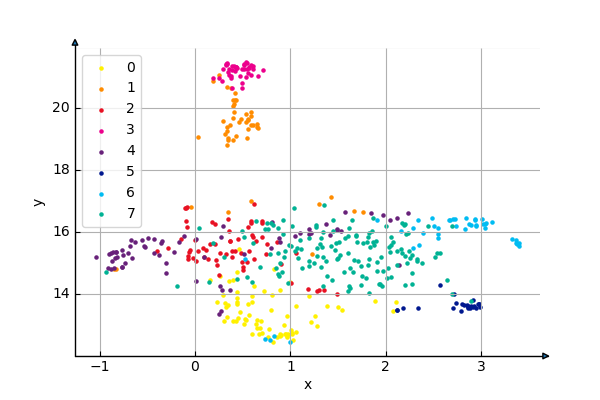}\\
(a) VGG $n=4$, $9$ clusters & (b) VGG $n=7$, $8$ clusters\\
\includegraphics[width=6.0cm]{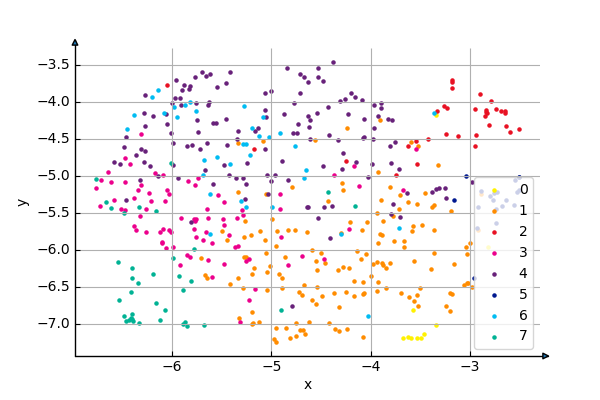} & \includegraphics[width=6.0cm]{scatter4_512images_no_bias_resnet_no_bias.png}\\
(c) VGG $n=14$, $8$ clusters & (d) ResNet $n=4$, $11$ clusters\\
\includegraphics[width=6.0cm]{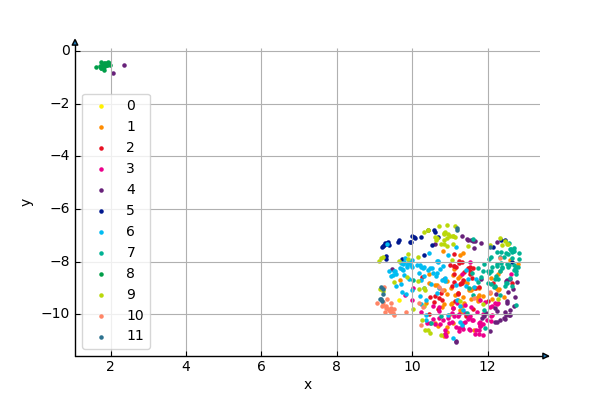} & \includegraphics[width=6.0cm]{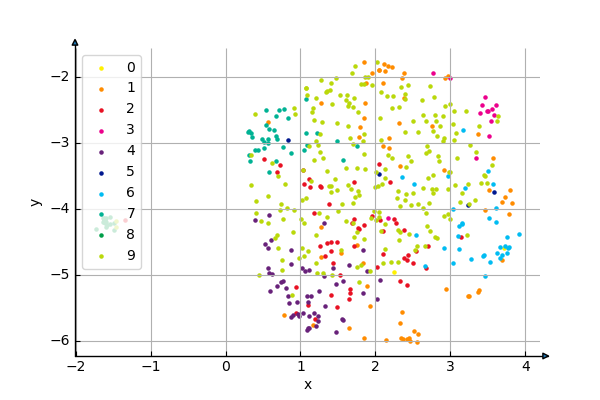}\\
(e) ResNet $n=7$, $12$ clusters & (f) ResNet $n=14$, $10$ clusters
\end{tabular}
\caption{Smallest values of $n$ seem to present denser clusters. Figures (a), (b), and (c) represent the scatter plots of feature maps of VGG16 while Figures (d), (e), and (f) represent those of ResNet according to the representation using $n \in \{4,7,14\}$ and $t=5$, respectively. The colors represent the clusters obtained by K-means. We used the value $k$ equal to $9,8$, and $8$ for VGG and, $11, 12$, and $10$ for ResNet  (for each patch size) chosen by the Elbow curve method from the Inertia presented in Figure~\ref{fig:silhouette_inertia}(b).}
\label{fig:scatter_clusters_supp}
\end{figure}

To test MAGE, we \revFour{varied} the representation patch size $\sizepatch \times \sizepatch$ with the following values: $2, 4, 7, 14, 28$ and $56$. The number $t$ of patches from each image to compose the representation \revFour{was} equal to $1,5$, and $10$. We \revFour{used} a subset of $512$ images, half from each class. For each configuration, we \revFour{applied} the $k$-means, and we \revFour{varied} the number of clusters $k \in [2,25[$. We \revFour{used} the metrics Silhouette and Inertia (distance of each sample to its cluster centroid). Inertia \revFour{was} used to choose the number $k$ of clusters.

Smaller sizes of patches present better quality (higher Silhouette); however, they fail to capture interpretable structures. To analyze fewer configurations, we \revFour{visualized} the dispersion (scatter plots) and central feature maps to each cluster using $n \in \{4,7,14\}$, $t=5$ (intermediate value of $t$ with smaller inertia than $t=10$). That choice removes the smallest $n$ value (representing less interpretable components), and the bigger $n$ values with less interesting Silhouette values. We selected $k$, for each configuration by using the Elbow curve method~\footnote{https://kneed.readthedocs.io} and Inertia.

We show the scatter plots that compare the spatial position of feature maps for patch sizes $n \in \{4,7,14\}$ in Fig.~\ref{fig:scatter_clusters_supp} for VGG16. The scatter plot visualizations show greater intra-cluster sparsity for larger patch sizes. It should be noted that for visualization purposes, we reduced the representation dimensionality using the UMAP technique~\cite{mcinnes:2018:umap}, with the parameters $n\_neighbors = 50$, $min\_dist = 0.0$, and Euclidean distance. Colors designate the different assigned clusters for each of them.

These results highlight that the size of the patches is a crucial parameter. Smaller patches can help to provide better concept clusters; however, they will probably capture fewer interpretable structures (the same xAI pixel-level technique's problem). Moreover, when using these smaller patches, the number of analyzed regions increases together with the number of computations. On the other hand, big patches will not cluster feature maps well enough, with poorer evaluation results and bigger sparsity (Fig.~\ref{fig:scatter_clusters_supp}). As the feature map cluster centers were reasonably similar, we continued the experiments with the $n=4$ clusters (small but not the smallest).

\newpage

\subsection{Relation between CaOC and Probabilities}
\label{ap:exp_caoc}

\begin{figure}[!ht]
\centering
\begin{tabular}{cccc}
\includegraphics[width=6.0cm]{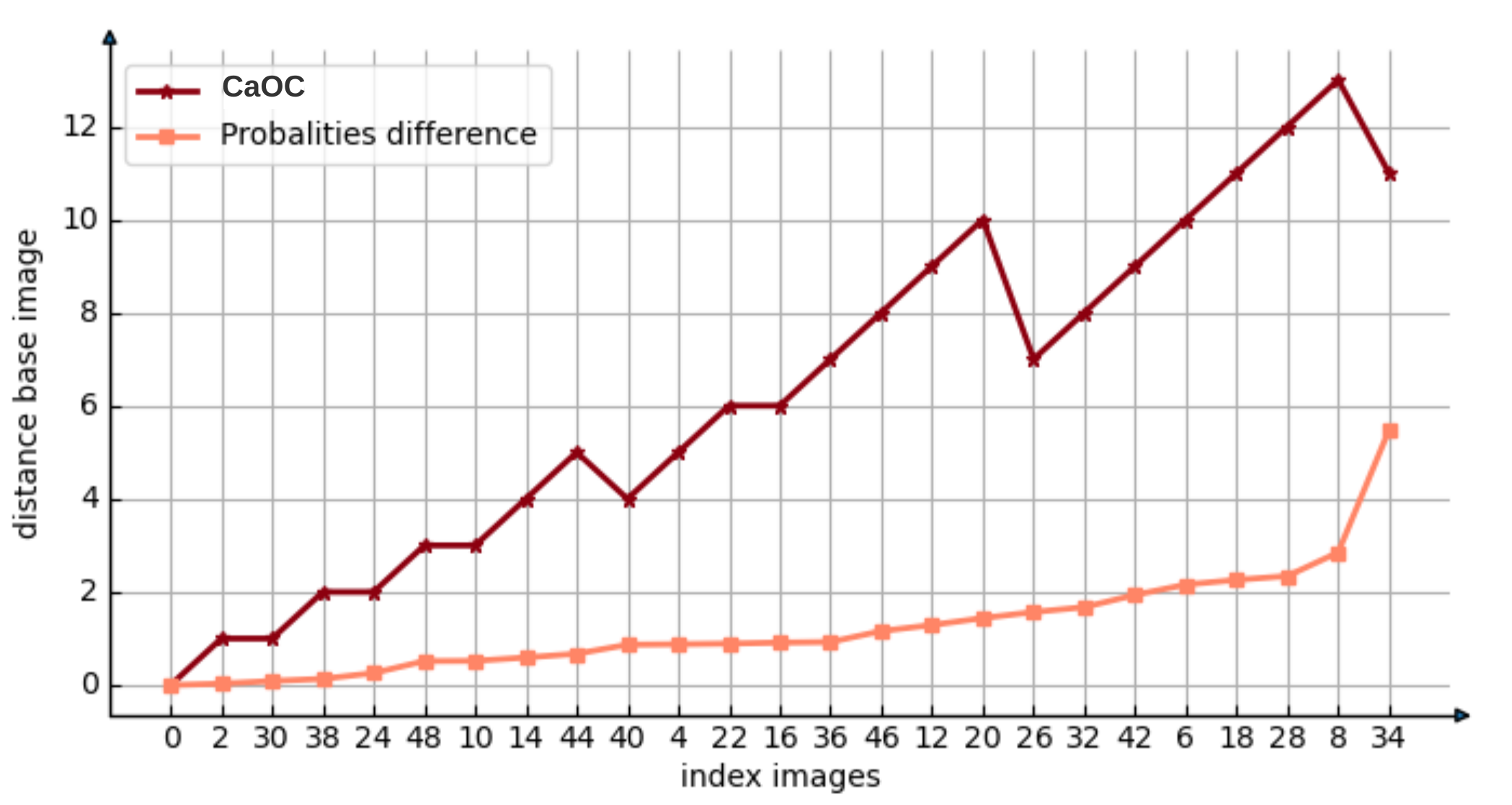}
&
\includegraphics[width=6.0cm]{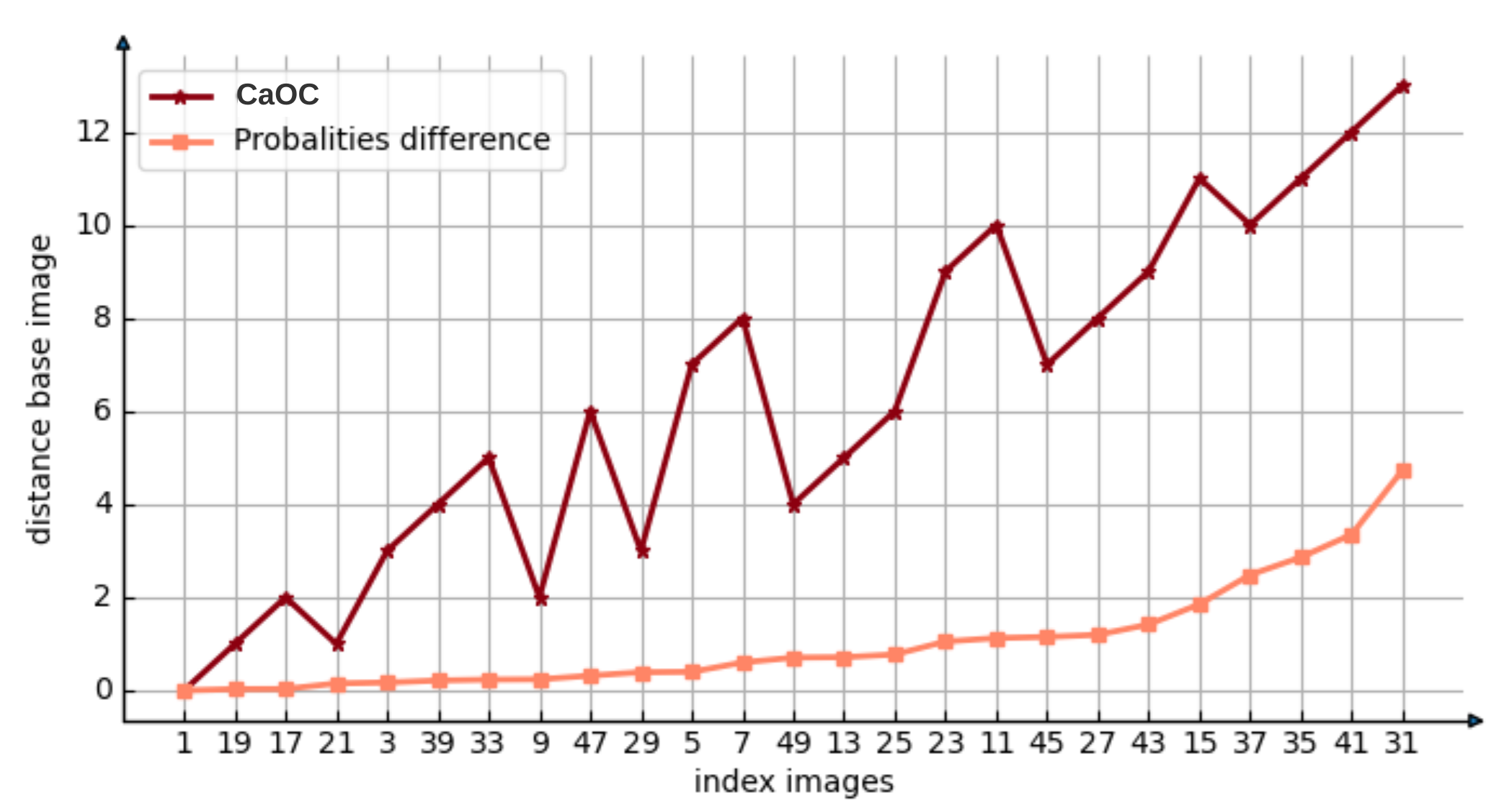}\\
(a) VGG -- dog & (b) VGG -- cat\\
\includegraphics[width=6.0cm]{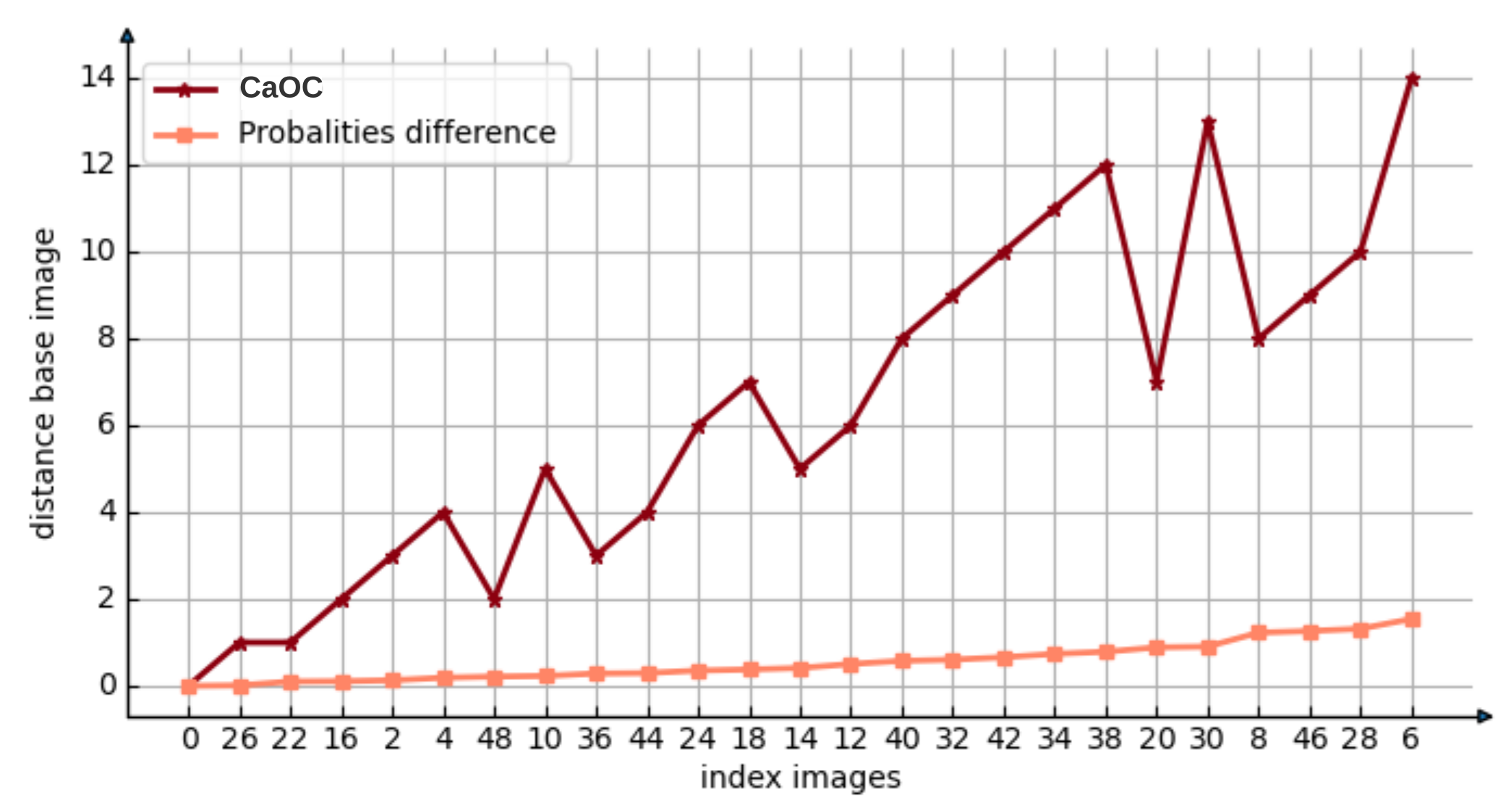}
& 
\includegraphics[width=6.0cm]{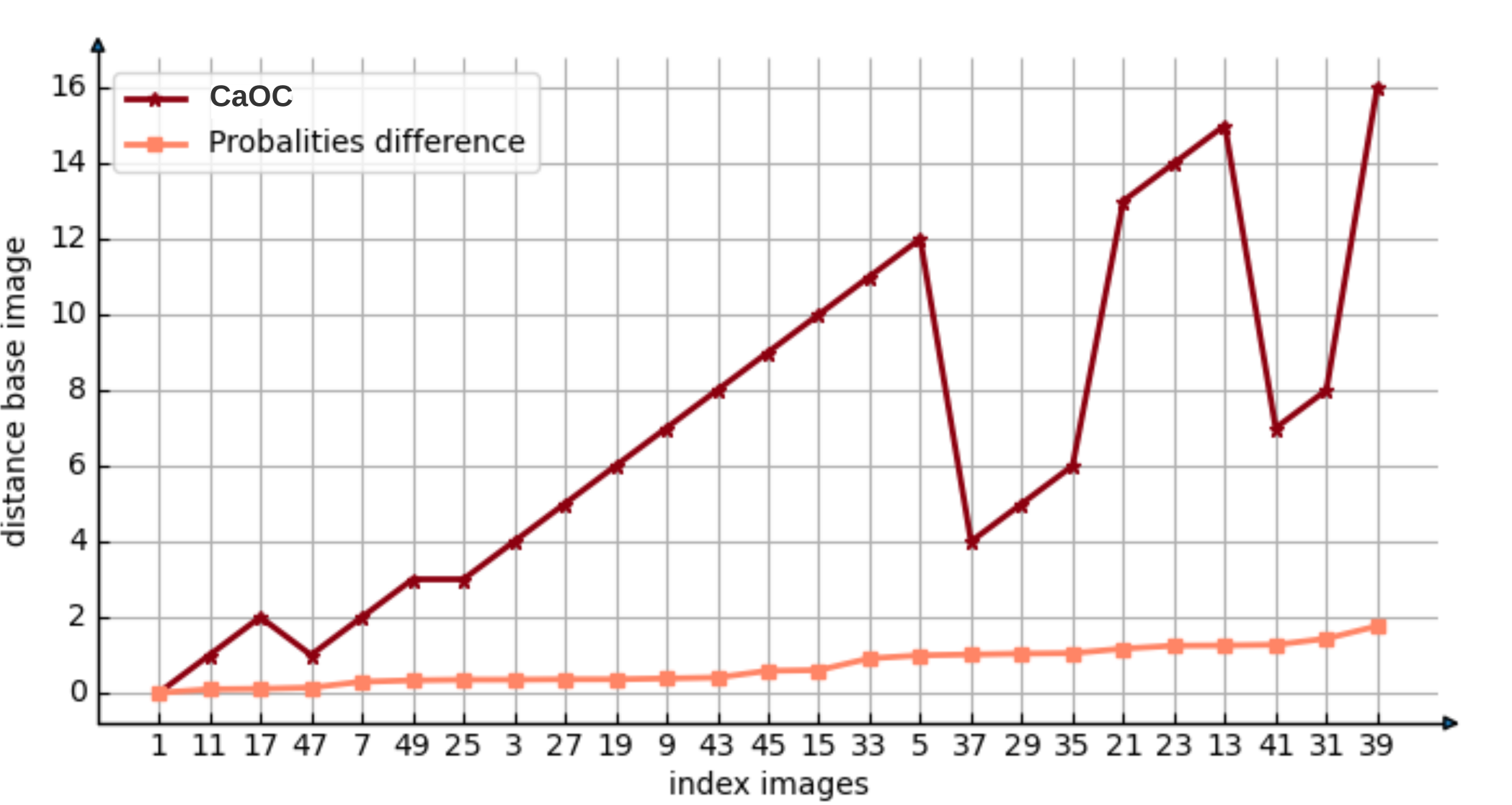}\\
(c) ResNet -- dog & (d) ResNet -- cat
\end{tabular}
\caption{\textit{CaOC} and \textit{Probabilities difference} behave differently. Based on a dog (index 0) and cat (index 1) image, we calculated the difference to 24 other images of each class, using \textit{CaOC} and \textit{Probabilities difference}. We ordered the images according to the distances obtained by \textit{Probabilities difference}. \textit{CaOC} presents discontinuities in the graph in relation to this order. Even indexes are dogs (Figures (a) and (c)), and odd indexes are cats (Figures (b) and (d)).}
\label{fig:comp_sample}
\end{figure}

\begin{figure}[!ht]
\centering
\begin{subfigure}[t]{0.44\textwidth}
\includegraphics[width=\linewidth]{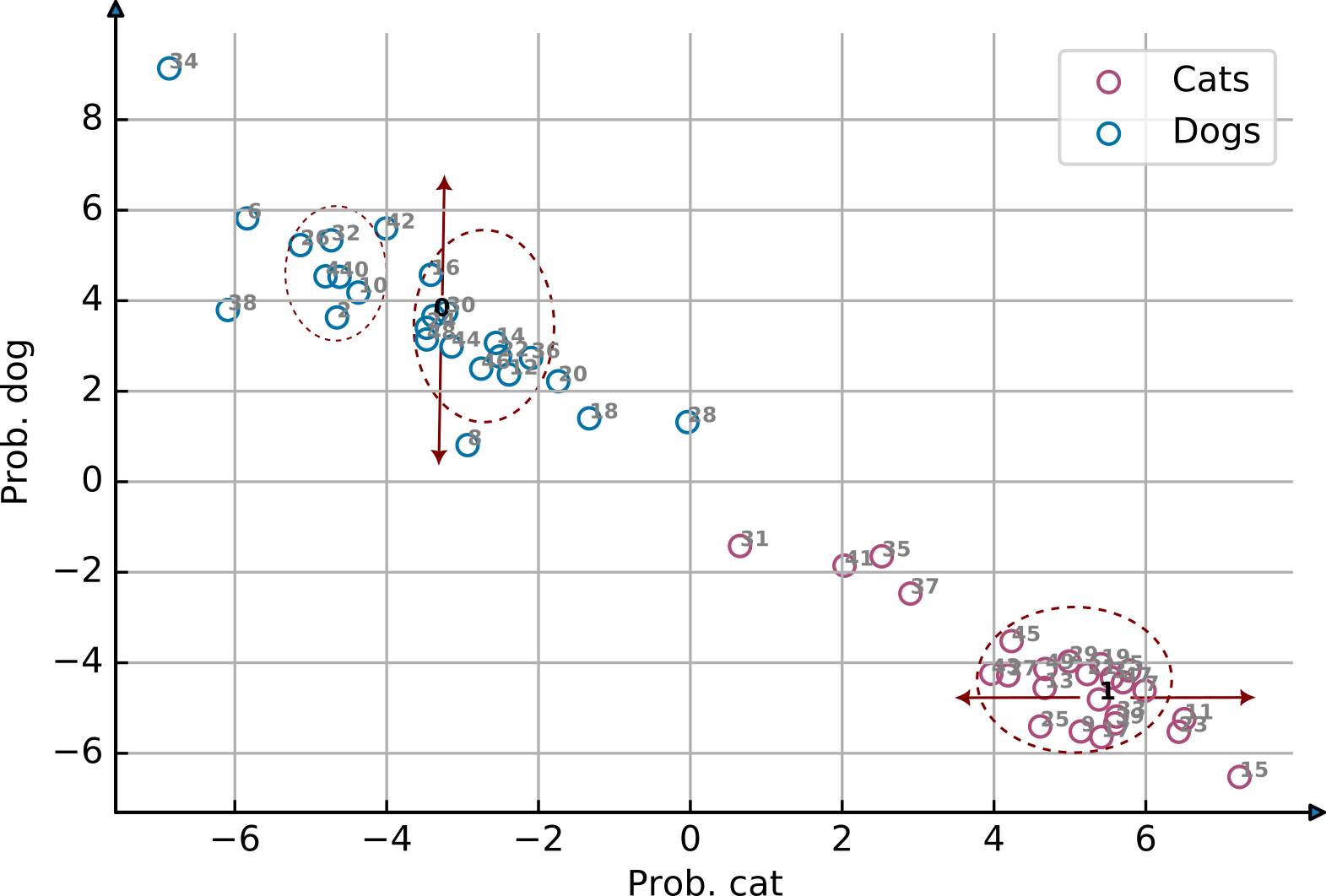}
\caption{VGG16 50 images}
\end{subfigure}~
\begin{subfigure}[t]{0.44\textwidth}
\includegraphics[width=\linewidth]{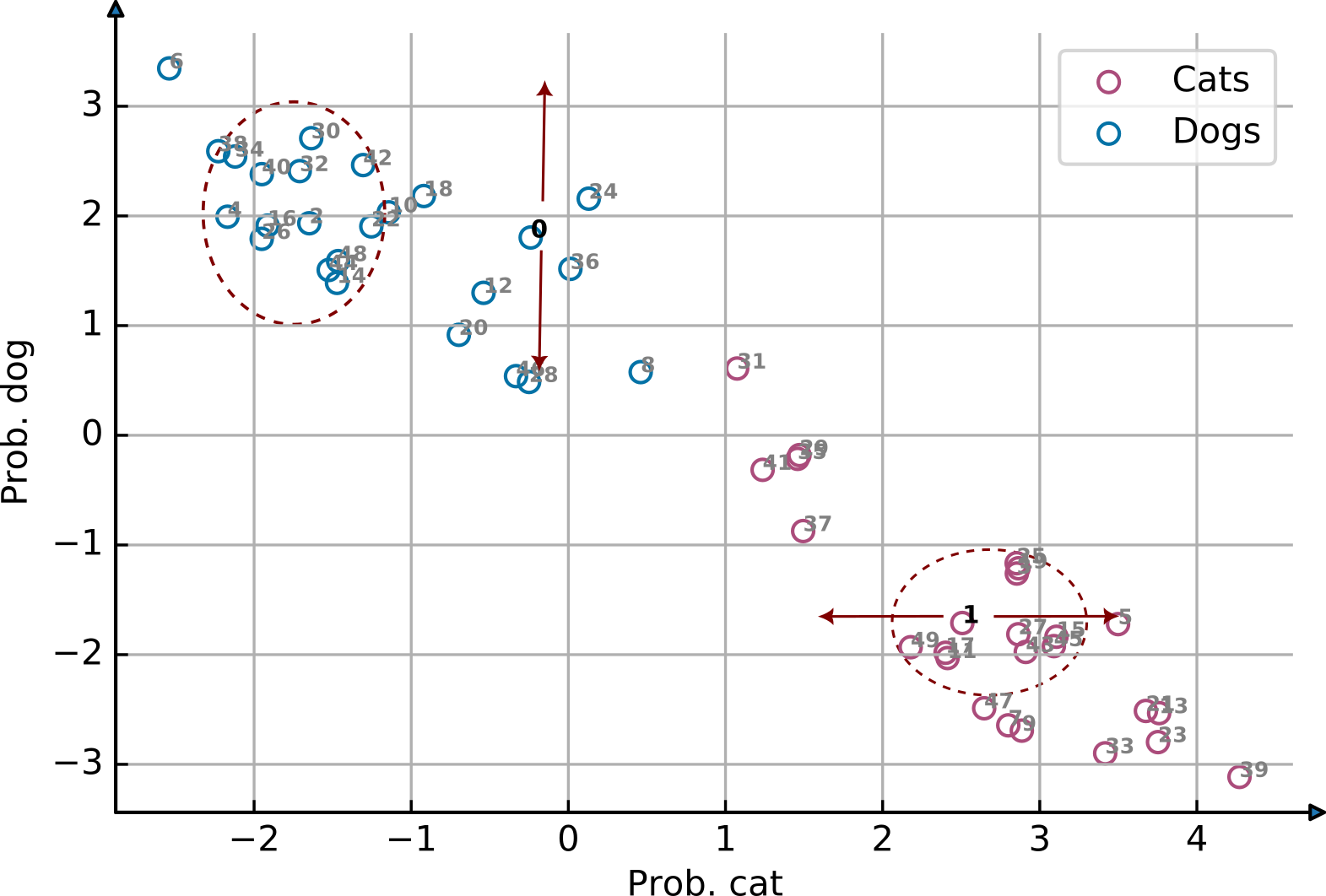}
\caption{ResNet18 50 images}
\end{subfigure}

\caption{Scatter plots of 50 images using VGG16/ResNet-18 final non-normalized probabilities. $\CAOC$ and $\PD$ are based on the dog's probability axis for dogs and the cat's probability axis for cats. Black numbers represent the base samples for the differences in Fig.~\ref{fig:comp_sample}. ResNet-18 is sparser.}
\label{fig:comp_sample_scatter}
\end{figure}

We show the difference between the $\CAOC$ metric and the \textit{probabilities difference} ($\PD$) used as a metric. We selected 50 images from the dataset (to make the visualization easier), 25 from each class (dogs as even numbers and cats as odd numbers), and we calculated the difference from image 0 (dog) and image 1 (cat) to all the others (from their respective classes) using $\CAOC$ and $\PD$. We present the results for the dog class in Fig.~\ref{fig:comp_sample}. In this figure, we ordered the images with respect to the distance obtained by $\PD$s to observe the behavior of $\CAOC$ as a function of $\PD$. As $\PD$ increases, $\CAOC$ does not follow a continuous behavior. Thus, we \revFour{closely looked at} discontinuities such as sample 34 (Fig.~\ref{fig:comp_sample}(a)) in  Fig~\ref{fig:comp_sample_scatter}. We \revFour{projected} 50 samples using their classes' (non-normalized) probabilities as coordinates (that is, the activations before the softmax). Samples from VGG~\ref{fig:comp_sample_scatter}(a) present a denser region close to sample 0 than to samples from ResNet18~\ref{fig:comp_sample_scatter}(b), which is reflected in Fig.~\ref{fig:comp_sample}, presenting more ResNet18 discontinuities. This density-awareness is expected from $\CAOC$. The sparsity represents fewer model-informative regions (based on the dataset), which thus count less for deciding on the model's globally important patterns.

\newpage

\subsection{Knowledge Discovering}
\label{ap:knowledge}

\textbf{Finding concepts:} We selected six MAG generated clusters from ResNet-18 and VGG16. We visualized each cluster through Ms-IV applied to 16 images (8 cats and 8 dogs) from the top-middle ranking positions. From a ranking of 512 images, we started at position 100 to avoid sparsity in higher and lower positions (possible outliers). We presented the Ms-IV visualizations of these image subsets to the research participants and asked which animal part corresponded to the lighter regions in dogs and cats. As we limited the analysis to six clusters per network, there were a total of 12 image subsets.

\begin{figure*}[!ht]
  \centering
  \includegraphics[width=11cm]{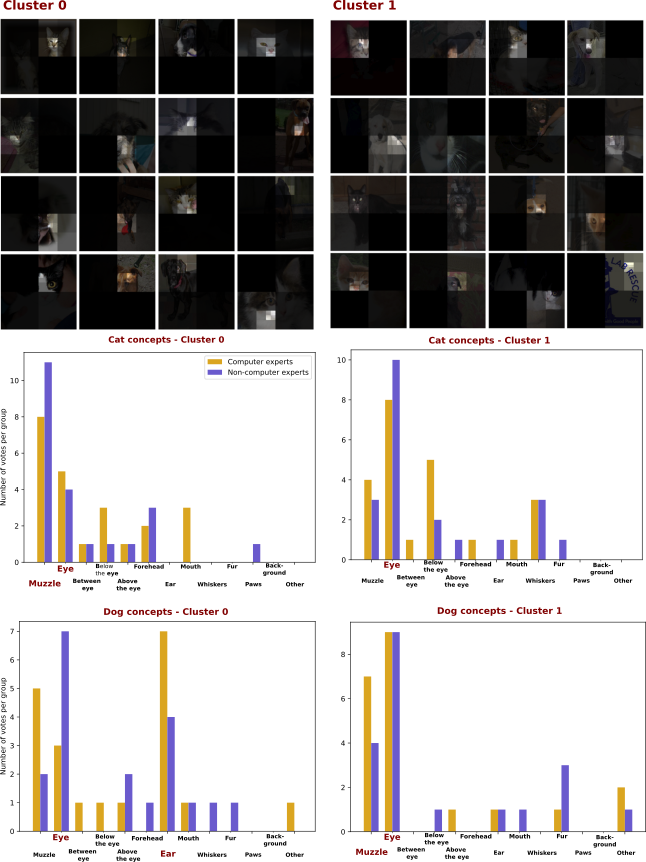}
  \caption{Visualizations obtained for clusters 0 and 1 of ResNet-18 and results of selected concepts, by 24 participants, to describe the two classes separately. According to the answers, cluster 0 presents the \textbf{eye} and \textbf{muzzle} of cats, while highlighting the \textbf{eye} and \textbf{ear} of dogs. Cluster 1 presents the \textbf{eye} for both classes.}
  \label{fig:kd_cluster01}
\end{figure*}

\begin{figure*}[!ht]
  \centering
  \includegraphics[width=11cm]{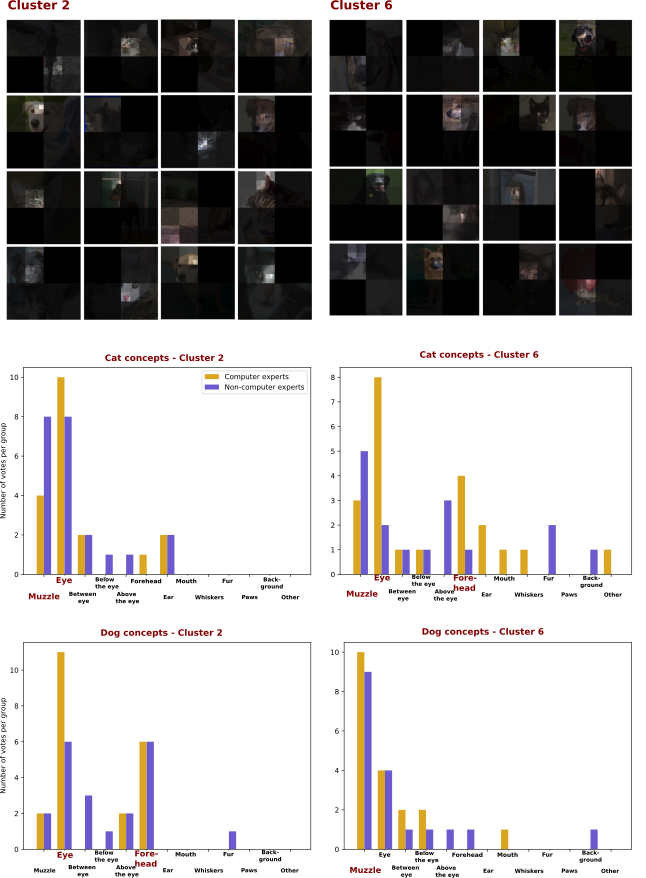}
  \caption{Visualizations obtained for clusters 2 and 6 of ResNet-18 and results of selected concepts, by 24 participants, to describe the two classes separately. According to the answers, cluster 2 presents the \textbf{eye} and \textbf{muzzle} of cats, while highlighting the \textbf{eye} and \textbf{forehead} of dogs. Cluster 6 presents the \textbf{muzzle} for dogs and a mix of concepts, \textbf{eye}, \textbf{muzzle}, and \textbf{forehead}, for cats.}
  \label{fig:kd_cluster26}
\end{figure*}

\begin{figure*}[!ht]
  \centering
  \includegraphics[width=11cm]{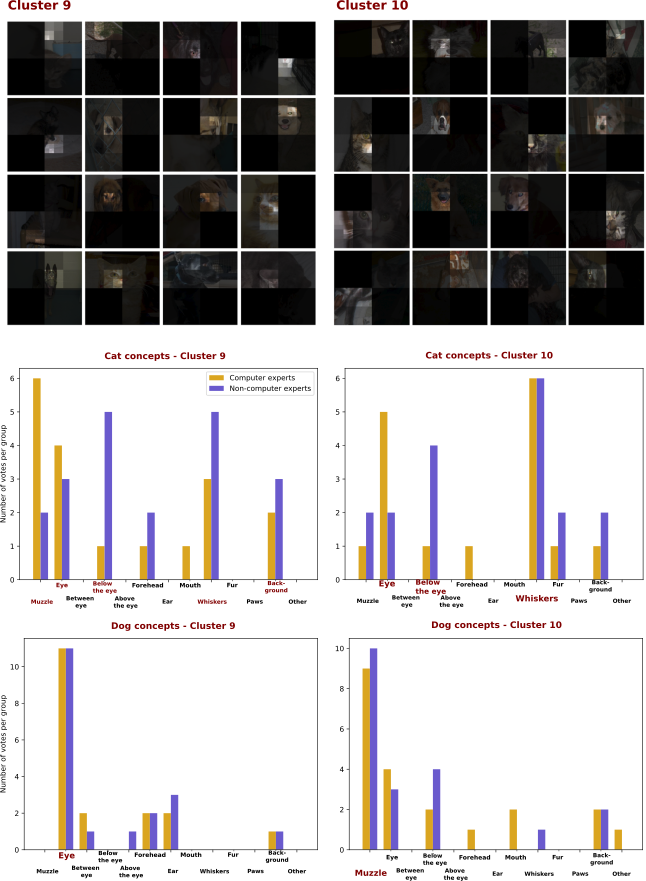}
  \caption{Visualizations obtained for clusters 9 and 10 of ResNet-18 and the results of selected concepts were described separately by 24 participants. According to the answers, cluster 9 seems not to be well-formed for the cat, but highlights the dog's \textbf{eye}. Cluster 10 presents the \textbf{muzzle} for dogs and the \textbf{eyes}, \textbf{below the eyes}, and \textbf{whiskers} for cats.}
  \label{fig:kd_cluster910}
\end{figure*}

\begin{figure*}[!ht]
  \centering
  \includegraphics[width=11cm]{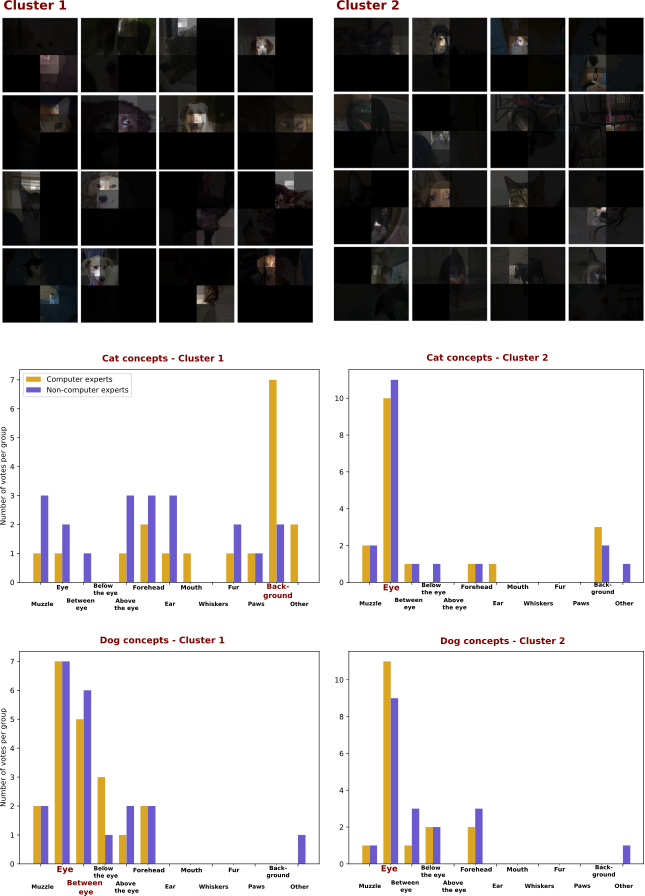}
  \caption{Visualizations obtained for clusters 1 and 2 of VGG16 and results of selected concepts, by 24 participants, to describe the two classes separately. According to the answers, cluster 1 seems not to detect cats well, highlighting the \textbf{background}, but highlights the dogs' \textbf{eyes} and the area \textbf{between eyes}. Cluster 2 presents the \textbf{eyes} for both animals.}
  \label{fig:kd_cluster12}
\end{figure*}

\begin{figure*}[!ht]
  \centering
  \includegraphics[width=11cm]{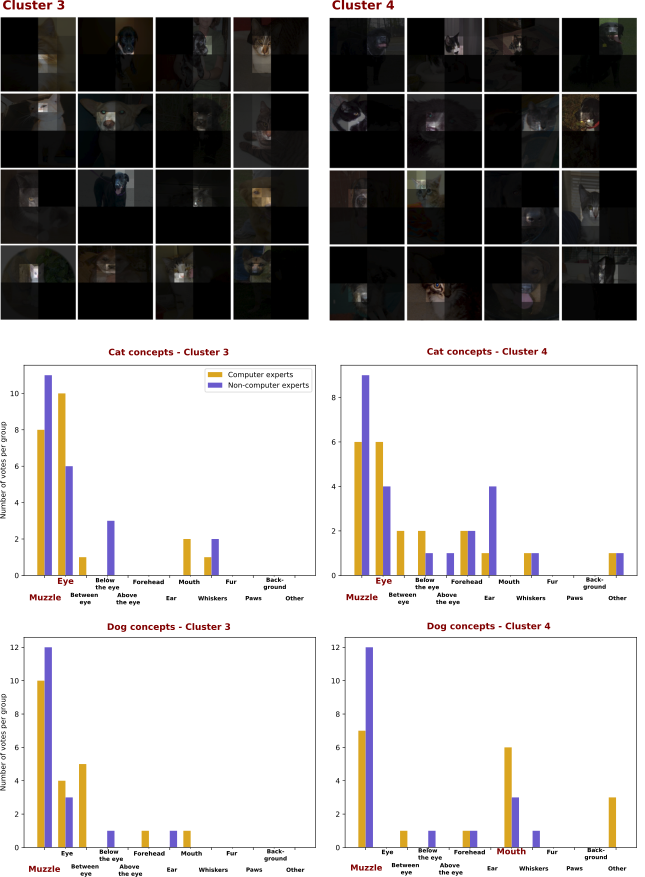}
  \caption{Visualizations obtained for clusters 3 and 4 of VGG16 and results of selected concepts, by 24 participants, to describe the two classes separately. According to the answers, cluster 3 seems not to detect the \textbf{muzzle} for both animals and the \textbf{eye} for cats. Cluster 4 presents also the \textbf{muzzle} and \textbf{eye} for cats, but the \textbf{muzzle} and \textbf{mouth} for dogs.}
  \label{fig:kd_cluster34}
\end{figure*}

\begin{figure*}[!ht]
  \centering
  \includegraphics[width=11cm]{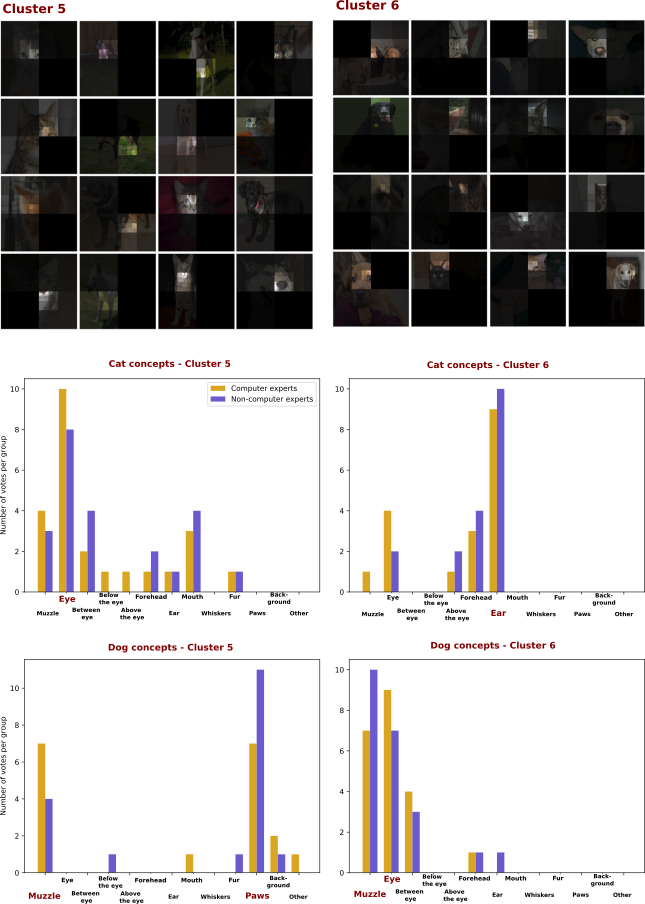}
  \caption{Visualizations obtained for clusters 5 and 6 of VGG16 and results of selected concepts, by 24 participants, to describe the two classes separately. According to the answers, cluster 5 seems not to detect the \textbf{eye} for cats and the \textbf{muzzle} and \textbf{paws} for dogs. Cluster 6 presents the \textbf{ear} of cats and the \textbf{muzzle} and \textbf{eye} for dogs.}
  \label{fig:kd_cluster56}
\end{figure*}

The 12 obtained subsets and answers are presented in Figures~\ref{fig:kd_cluster01}, \ref{fig:kd_cluster26}, and \ref{fig:kd_cluster910} for ResNet-18 and in Figures~\ref{fig:kd_cluster12}, \ref{fig:kd_cluster34}, and \ref{fig:kd_cluster56} for VGG16.

In general, out of the 13 proposed concepts, fewer than three of them received most of the participants' votes for each cluster. There was agreement about concepts for both computer and non-computer experts. Concepts such as \textbf{eyes} and \textbf{muzzle} were the most observed. %

\end{document}